\documentclass[10pt,twocolumn,letterpaper]{article}

\usepackage{cvpr}
\usepackage{times}
\usepackage{epsfig}
\usepackage{graphicx}
\usepackage{amsmath}
\usepackage{amssymb}
\usepackage{nicefrac}
\usepackage{mathtools}
\usepackage{adjustbox}

\usepackage{caption}
\usepackage{tikz}
\usepackage{pgfplots}
\usetikzlibrary{spy,calc}
\usepackage{array}
\newcolumntype{x}[1]{>{\centering\arraybackslash\hspace{0pt}}p{#1}}

\newif\ifblackandwhitecycle
\gdef\patternnumber{0} 

\pgfkeys{/tikz/.cd,
    zoombox paths/.style={
        draw=orange,
        very thick
    },
    black and white/.is choice,
    black and white/.default=static,
    black and white/static/.style={ 
        draw=white,   
        zoombox paths/.append style={
            draw=white,
            postaction={
                draw=black,
                loosely dashed
            }
        }
    },
    black and white/static/.code={
        \gdef\patternnumber{1}
    },
    black and white/cycle/.code={
        \blackandwhitecycletrue
        \gdef\patternnumber{1}
    },
    black and white pattern/.is choice,
    black and white pattern/0/.style={},
    black and white pattern/1/.style={    
            draw=white,
            postaction={
                draw=black,
                dash pattern=on 2pt off 2pt
            }
    },
    black and white pattern/2/.style={    
            draw=white,
            postaction={
                draw=black,
                dash pattern=on 4pt off 4pt
            }
    },
    black and white pattern/3/.style={    
            draw=white,
            postaction={
                draw=black,
                dash pattern=on 4pt off 4pt on 1pt off 4pt
            }
    },
    black and white pattern/4/.style={    
            draw=white,
            postaction={
                draw=black,
                dash pattern=on 4pt off 2pt on 2 pt off 2pt on 2 pt off 2pt
            }
    },
    zoomboxarray inner gap/.initial=5pt,
    zoomboxarray columns/.initial=2,
    zoomboxarray rows/.initial=2,
    subfigurename/.initial={},
    figurename/.initial={zoombox},
    zoomboxarray/.style={
        execute at begin picture={
            \begin{scope}[
                spy using outlines={%
                    zoombox paths,
                    width=\imagewidth / \pgfkeysvalueof{/tikz/zoomboxarray columns} - (\pgfkeysvalueof{/tikz/zoomboxarray columns} - 1) / \pgfkeysvalueof{/tikz/zoomboxarray columns} * \pgfkeysvalueof{/tikz/zoomboxarray inner gap} -\pgflinewidth,
                    height=\imageheight / \pgfkeysvalueof{/tikz/zoomboxarray rows} - (\pgfkeysvalueof{/tikz/zoomboxarray rows} - 1) / \pgfkeysvalueof{/tikz/zoomboxarray rows} * \pgfkeysvalueof{/tikz/zoomboxarray inner gap}-\pgflinewidth,
                    magnification=3,
                    every spy on node/.style={
                        zoombox paths
                    },
                    every spy in node/.style={
                        zoombox paths
                    }
                }
            ]
        },
        execute at end picture={
            \end{scope}
     \gdef\patternnumber{0}
        },
        spymargin/.initial=0.5em,
        zoomboxes xshift/.initial=1,
        zoomboxes right/.code=\pgfkeys{/tikz/zoomboxes xshift=1},
        zoomboxes left/.code=\pgfkeys{/tikz/zoomboxes xshift=-1},
        zoomboxes yshift/.initial=0,
        zoomboxes above/.code={
            \pgfkeys{/tikz/zoomboxes yshift=1},
            \pgfkeys{/tikz/zoomboxes xshift=0}
        },
        zoomboxes below/.code={
            \pgfkeys{/tikz/zoomboxes yshift=-1},
            \pgfkeys{/tikz/zoomboxes xshift=0}
        },
        caption margin/.initial=0.4ex,
    },
    adjust caption spacing/.code={},
    image container/.style={
        inner sep=0pt,
        at=(image.north),
        anchor=north,
        adjust caption spacing
    },
    zoomboxes container/.style={
        inner sep=0pt,
        at=(image.north),
        anchor=north,
        name=zoomboxes container,
        xshift=\pgfkeysvalueof{/tikz/zoomboxes xshift}*(\imagewidth+\pgfkeysvalueof{/tikz/spymargin}),
        yshift=\pgfkeysvalueof{/tikz/zoomboxes yshift}*(\imageheight+\pgfkeysvalueof{/tikz/spymargin}+\pgfkeysvalueof{/tikz/caption margin}),
        adjust caption spacing
    },
    calculate dimensions/.code={
        \pgfpointdiff{\pgfpointanchor{image}{south west} }{\pgfpointanchor{image}{north east} }
        \pgfgetlastxy{\imagewidth}{\imageheight}
        \global\let\imagewidth=\imagewidth
        \global\let\imageheight=\imageheight
        \gdef\columncount{1}
        \gdef\rowcount{1}
        
    },
    image node/.style={
        inner sep=0pt,
        name=image,
        anchor=south west,
        append after command={
            [calculate dimensions]
            node [image container,subfigurename=\pgfkeysvalueof{/tikz/figurename}-image] {\phantomimage}
            node [zoomboxes container,subfigurename=\pgfkeysvalueof{/tikz/figurename}-zoom] {\phantomimage}
        }
    },
    color code/.style={
        zoombox paths/.append style={draw=#1}
    },
    connect zoomboxes/.style={
    spy connection path={\draw[draw=none,zoombox paths] (tikzspyonnode) -- (tikzspyinnode);}
    },
    help grid code/.code={
        \begin{scope}[
                x={(image.south east)},
                y={(image.north west)},
                font=\footnotesize,
                help lines,
                overlay
            ]
            \foreach \x in {0,1,...,9} { 
                \draw(\x/10,0) -- (\x/10,1);
                \node [anchor=north] at (\x/10,0) {0.\x};
            }
            \foreach \y in {0,1,...,9} {
                \draw(0,\y/10) -- (1,\y/10);                        \node [anchor=east] at (0,\y/10) {0.\y};
            }
        \end{scope}    
    },
    help grid/.style={
        append after command={
            [help grid code]
        }
    },
}

\newcommand\phantomimage{%
    \phantom{%
        \rule{\imagewidth}{\imageheight}%
    }%
}
\newcommand\zoombox[2][]{
    \begin{scope}[zoombox paths]
        \pgfmathsetmacro\xpos{
            (\columncount-1)*(\imagewidth / \pgfkeysvalueof{/tikz/zoomboxarray columns} + \pgfkeysvalueof{/tikz/zoomboxarray inner gap} / \pgfkeysvalueof{/tikz/zoomboxarray columns} ) + \pgflinewidth
        }
        \pgfmathsetmacro\ypos{
            (\rowcount-1)*( \imageheight / \pgfkeysvalueof{/tikz/zoomboxarray rows} + \pgfkeysvalueof{/tikz/zoomboxarray inner gap} / \pgfkeysvalueof{/tikz/zoomboxarray rows} ) + 0.5*\pgflinewidth
        }
        \edef\dospy{\noexpand\spy [
            #1,
            zoombox paths/.append style={
                black and white pattern=\patternnumber
            },
            every spy on node/.append style={#1},
            x=\imagewidth,
            y=\imageheight
        ] on (#2) in node [anchor=north west] at ($(zoomboxes container.north west)+(\xpos pt,-\ypos pt)$);}
        \dospy
        \pgfmathtruncatemacro\pgfmathresult{ifthenelse(\columncount==\pgfkeysvalueof{/tikz/zoomboxarray columns},\rowcount+1,\rowcount)}
        \global\let\rowcount=\pgfmathresult
        \pgfmathtruncatemacro\pgfmathresult{ifthenelse(\columncount==\pgfkeysvalueof{/tikz/zoomboxarray columns},1,\columncount+1)}
        \global\let\columncount=\pgfmathresult
        \ifblackandwhitecycle
            \pgfmathtruncatemacro{\newpatternnumber}{\patternnumber+1}
            \global\edef\patternnumber{\newpatternnumber}
        \fi
    \end{scope}
}

\usepackage[acronym,shortcuts]{glossaries}

\usepackage{setspace}

\usepackage[pagebackref=true,breaklinks=true,letterpaper=true,colorlinks,bookmarks=false]{hyperref}

\usepackage{booktabs}

\cvprfinalcopy 

\def\cvprPaperID{1948} 

\ifcvprfinal\fi

\makeglossaries

\newacronym{SISR}{SISR}{single image super-resolution}
\newacronym{PSNR}{PSNR}{peak signal-to-noise ratio}
\newacronym{MSE}{MSE}{mean squared error}
\newacronym{CNN}{CNN}{convolutional neural network}
\newacronym{ESPCN}{ESPCN}{efficient sub-pixel convolutional neural network}
\newacronym{DRCN}{DRCN}{deeply-recursive convolutional network}
\newacronym{ResNet}{ResNet}{residual network}
\newacronym{GAN}{GAN}{generative adversarial network}
\newacronym{LR}{LR}{low-resolution}
\newacronym{HR}{HR}{high-resolution}
\newacronym{SR}{SR}{super-resolution}
\newacronym{MOS}{MOS}{mean opinion score}
\newacronym{SSIM}{SSIM}{structural similarity}
\newacronym{PSNR-HVS}{PSNR-HVS}{peak signal-to-noise ratio - human visual system}
\newacronym{MS-SSIM}{MS-SSIM}{multi-scale structural similarity}
\newacronym{SRGAN}{SRGAN}{super-resolution generative adversarial network}

\usepackage{subcaption}

\begin{document}

\title{Photo-Realistic Single Image Super-Resolution Using a Generative Adversarial Network}

\onehalfspacing
\author{Christian Ledig, Lucas Theis, Ferenc Husz\'{a}r, Jose Caballero, Andrew Cunningham, \\Alejandro Acosta, Andrew Aitken, Alykhan Tejani, Johannes Totz, Zehan Wang, Wenzhe Shi\\
        Twitter\\
        {\tt\footnotesize \{cledig,ltheis,fhuszar,jcaballero,aacostadiaz,aaitken,atejani,jtotz,zehanw,wshi\}@twitter.com}
        }

\maketitle

\singlespacing
\begin{abstract}
    Despite the breakthroughs in accuracy and speed of single image super-resolution using faster and deeper convolutional neural networks, one central problem remains largely unsolved: how do we recover the finer texture details when we super-resolve at large upscaling factors?
    The behavior of optimization-based super-resolution methods is principally driven by the choice of the objective function. Recent work has largely focused on minimizing the mean squared reconstruction error. The resulting estimates have high peak signal-to-noise ratios, but they are often lacking high-frequency details and are perceptually unsatisfying in the sense that they fail to match the fidelity expected at the higher resolution.
  In this paper, we present SRGAN, a generative adversarial network (GAN) for image super-resolution (SR). To our knowledge, it is the first framework capable of inferring photo-realistic natural images for $4\times$ upscaling factors. To achieve this, we propose a perceptual loss function which consists of an adversarial loss and a content loss. The adversarial loss pushes our solution to the natural image manifold using a discriminator network that is trained to differentiate between the super-resolved images and original photo-realistic images. In addition, we use a content loss motivated by perceptual similarity instead of similarity in pixel space. Our deep residual network is able to recover photo-realistic textures from heavily downsampled images on public benchmarks.
An extensive mean-opinion-score (MOS) test shows hugely significant gains in perceptual quality using SRGAN. The MOS scores obtained with SRGAN are closer to those of the original high-resolution images than to those obtained with any state-of-the-art method.
\end{abstract}

\section{Introduction}
The highly challenging task of estimating a \ac{HR} image from its \ac{LR} counterpart is referred to as \ac{SR}.
	\ac{SR} received substantial attention from within the computer vision research community and has a wide range of applications \cite{yang2007spatial,Zou12,Nasrollahi2014}.

\begin{figure}[ht] 
  	\begin{tabular}{cc}
  		$4\times$ SRGAN (proposed) & original \\
     	\includegraphics[trim=0 0 0 0, clip, width=1.5in]{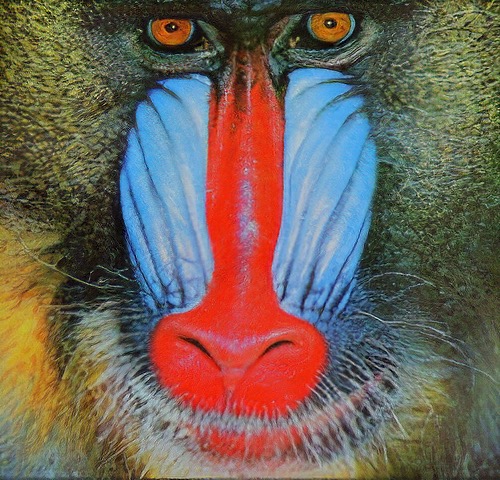} &
     	\includegraphics[trim=0 0 0 0, clip, width=1.5in]{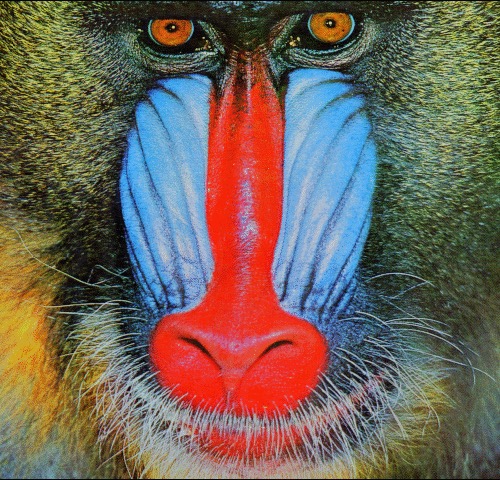} \\
     	
  	\end{tabular}
	\caption{Super-resolved image (left) is almost indistinguishable from original (right). [$4\times$ upscaling]} 
	\label{fig:exampleIntroFirst}
\end{figure}

	The ill-posed nature of the underdetermined \ac{SR} problem is particularly pronounced for high upscaling factors, for which texture detail in the reconstructed \ac{SR} images is typically absent.
	%
%
\begin{figure*}[ht] 
  	\begin{tabular}{cccc}
  		 bicubic &  SRResNet  & SRGAN & original\\
   		 (21.59dB/0.6423) &   (23.53dB/0.7832) & (21.15dB/0.6868) & \\
     	\includegraphics[trim=0pt 0pt 0pt 0pt, clip, width=1.55in]{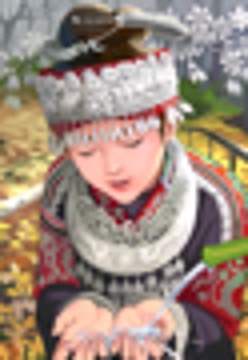} &
     	\includegraphics[trim=0pt 0pt 0pt 0pt, clip, width=1.55in]{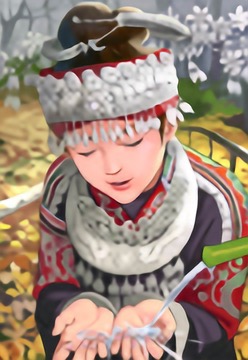} &
     	\includegraphics[trim=0pt 0pt 0pt 0pt, clip, width=1.55in]{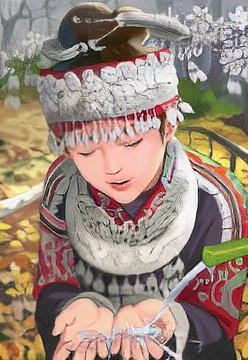} &
     	\includegraphics[trim=0pt 0pt 0pt 0pt, clip, width=1.55in]{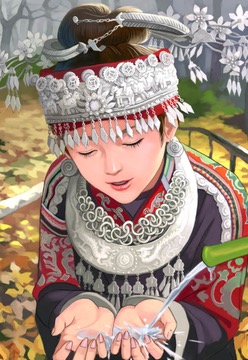} \\   	
  	\end{tabular}
	\caption{From left to right: bicubic interpolation, deep residual network optimized for MSE, deep residual generative adversarial network optimized for a loss more sensitive to human perception, original HR image. Corresponding PSNR and SSIM are shown in brackets. [$4\times$ upscaling]} 
	\label{fig:exampleIntro}
\end{figure*}
	The optimization target of supervised \ac{SR} algorithms is commonly the minimization of the \ac{MSE} between the recovered \ac{HR} image and the ground truth. This is convenient as minimizing \ac{MSE} also maximizes the \ac{PSNR}, which is a common measure used to evaluate and compare \ac{SR} algorithms \cite{Yang14benchmark}.
	 However, the ability of MSE (and PSNR) to capture perceptually relevant differences, such as high texture detail, is very limited as they are defined based on pixel-wise image differences \cite{Wang2003,Wang2004,Gupta2011}. This is illustrated in \figurename~\ref{fig:exampleIntro}, where highest PSNR does not necessarily reflect the perceptually better \ac{SR} result. The perceptual difference between the super-resolved and original image means that the recovered image is not photo-realistic as defined by Ferwerda \cite{Ferwerda2003}.

	In this work we propose a \ac{SRGAN} for which we employ a deep \ac{ResNet} with skip-connection and diverge from \ac{MSE} as the sole optimization target. Different from previous works, we define a novel perceptual loss using high-level feature maps of the VGG network \cite{simonyan2014very,Johnson16PercepLoss,bruna2016super} combined with a discriminator that encourages solutions perceptually hard to distinguish from the \ac{HR} reference images. 
	An example photo-realistic image that was super-resolved with a $4\times$ upscaling factor is shown in \figurename~\ref{fig:exampleIntroFirst}.
	
\subsection{Related work}
\subsubsection{Image super-resolution}
	Recent overview articles on image \ac{SR} include Nasrollahi and Moeslund \cite{Nasrollahi2014} or Yang et al. \cite{Yang14benchmark}. Here we will focus on \ac{SISR} and will not further discuss approaches that recover \ac{HR} images from multiple images \cite{Borman1998aSurvey,Farsiu2004}.
	
	Prediction-based methods were among the first methods to tackle \ac{SISR}. While these filtering approaches, \eg linear, bicubic or Lanczos \cite{Duchon1979} filtering, can be very fast, they oversimplify the \ac{SISR} problem and usually yield solutions with overly smooth textures.
	Methods that put particularly focus on edge-preservation have been proposed \cite{Allebach96, Li2001}.
	
	More powerful approaches aim to establish a complex mapping between low- and high-resolution image information and usually rely on training data. 
	Many methods that are based on example-pairs rely on \ac{LR} training patches for which the corresponding \ac{HR} counterparts are known. Early work was presented by Freeman et al. \cite{Freeman2000,Freeman2002}.
	Related approaches to the \ac{SR} problem originate in compressed sensing \cite{Yang08, Dong2011, zeyde2012single}.
	In Glasner et al. \cite{glasner2009super} the authors exploit patch redundancies across scales within the image to drive the \ac{SR}. This paradigm of self-similarity is also employed in Huang et al. \cite{Huang15selfexemplars}, where self dictionaries are extended by further allowing for small transformations and shape variations.
	Gu et al. \cite{gu2015convolutional} proposed a convolutional sparse coding approach that improves consistency by processing the whole image rather than overlapping patches.
	
	To reconstruct realistic texture detail while avoiding edge artifacts, Tai et al. \cite{Tai2010} combine an edge-directed \ac{SR} algorithm based on a gradient profile prior \cite{Sun2008} with the benefits of learning-based detail synthesis. Zhang et al. \cite{zhang2012multi} propose a multi-scale dictionary to capture redundancies of similar image patches at different scales. 
	To super-resolve landmark images, Yue et al. \cite{Yue2013} retrieve correlating \ac{HR} images with similar content from the web and propose a structure-aware matching criterion for alignment.
	
	Neighborhood embedding approaches upsample a \ac{LR} image patch by finding similar \ac{LR} training patches in a low dimensional manifold and combining their corresponding \ac{HR} patches for reconstruction \cite{timofte2013anchored,timofte2014a+}.
	In Kim and Kwon \cite{Kim10kernelregression} the authors emphasize the tendency of neighborhood approaches to overfit and formulate a more general map of example pairs using kernel ridge regression.
	The regression problem can also be solved with Gaussian process regression \cite{he2011single}, trees \cite{salvador2015naive} or Random Forests \cite{schulter2015fast}.
	In Dai et al. \cite{dai2015jointly} a multitude of patch-specific regressors is learned and the most appropriate regressors selected during testing. 
 
 	Recently \ac{CNN} based \ac{SR} algorithms have shown excellent performance.
 	In Wang et al. \cite{Wang2015} the authors encode a sparse representation prior into their feed-forward network architecture based on the learned iterative shrinkage and thresholding algorithm (LISTA) \cite{gregor2010learning}.
 	Dong et al. \cite{dong2014learning,dong2016image} used bicubic interpolation to upscale an input image and trained a three layer deep fully convolutional network end-to-end to achieve state-of-the-art \ac{SR} performance.
	Subsequently, it was shown that enabling the network to learn the upscaling filters directly can further increase performance both in terms of accuracy and speed \cite{dong2016accelerating,Shi2016ESPCN,Wang2016}. 
 	With their \ac{DRCN}, Kim et al. \cite{kim2016deeply} presented a highly performant architecture that allows for long-range pixel dependencies while keeping the number of model parameters small.
 	Of particular relevance for our paper are the works by Johnson et al. \cite{Johnson16PercepLoss} and Bruna et al. \cite{bruna2016super}, who rely on a loss function closer to perceptual similarity to recover visually more convincing \ac{HR} images.

\subsubsection{Design of convolutional neural networks}
	The state of the art for many computer vision problems is meanwhile set by specifically designed \ac{CNN} architectures following the success of the work by Krizhevsky et al. \cite{krizhevsky2012imagenet}. 
	
	It was shown that deeper network architectures can be difficult to train but have the potential to substantially increase the network's accuracy as they allow modeling mappings of very high complexity \cite{simonyan2014very,szegedy2015going}. To efficiently train these deeper network architectures, batch-normalization \cite{Ioffe2015} is often used to counteract the internal co-variate shift.
	Deeper network architectures have also been shown to increase performance for \ac{SISR}, \eg Kim et al. \cite{kim2016deeply} formulate a recursive \ac{CNN} and present state-of-the-art results.
	Another powerful design choice that eases the training of deep \ac{CNN}s is the recently introduced concept of residual blocks \cite{he2015deep} and skip-connections \cite{he2016identity,kim2016deeply}. Skip-connections relieve the network architecture of modeling the identity mapping that is trivial in nature, however, potentially non-trivial to represent with convolutional kernels.
	
	In the context of \ac{SISR} it was also shown that learning upscaling filters is beneficial in terms of accuracy and speed \cite{dong2016accelerating,Shi2016ESPCN,Wang2016}.
	This is an improvement over Dong et al. \cite{dong2016image} where bicubic interpolation is employed to upscale the LR observation before feeding the image to the \ac{CNN}. 

\begin{figure}[ht] 
     	\includegraphics[trim=0 0 0 0, clip, width=3.1in]{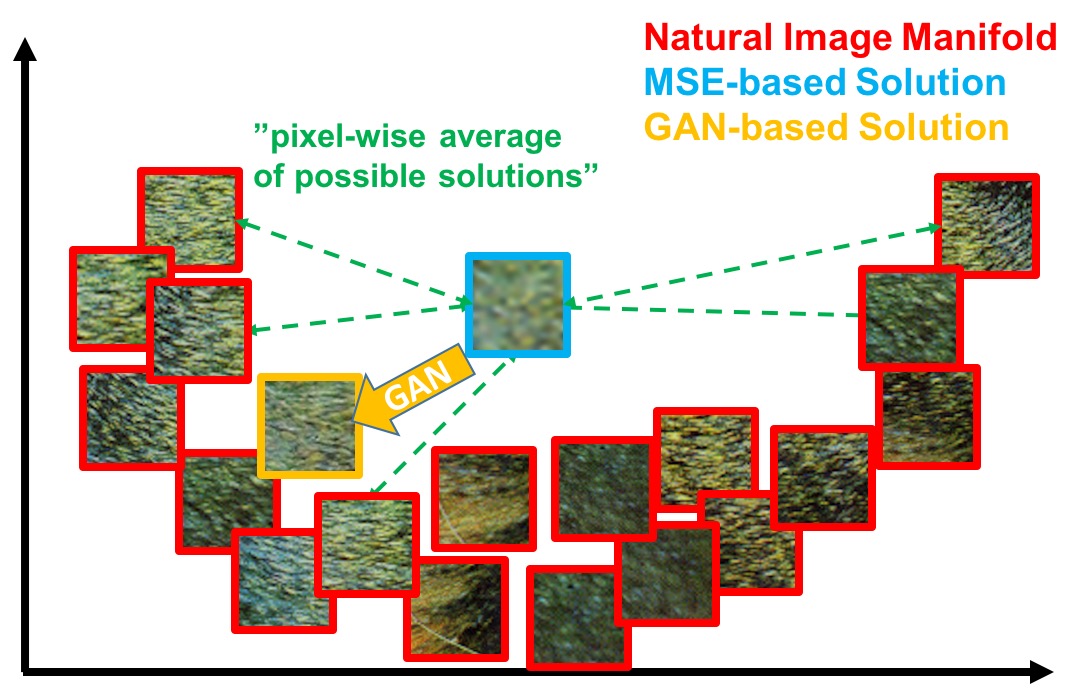} 
	\caption{Illustration of patches from the natural image manifold (red) and super-resolved patches obtained with MSE (blue) and GAN (orange). The MSE-based solution appears overly smooth due to the pixel-wise average of possible solutions in the pixel space, while GAN drives the reconstruction towards the natural image manifold producing perceptually more convincing solutions.} 
	\label{fig:manifold}
\end{figure}

\subsubsection{Loss functions}
Pixel-wise loss functions such as \ac{MSE} struggle to handle the uncertainty inherent in recovering lost high-frequency details such as texture: minimizing \ac{MSE} encourages finding pixel-wise averages of plausible solutions which are typically overly-smooth and thus have poor perceptual quality \cite{Mathieu2015, Johnson16PercepLoss, dosovitskiy2016generating, bruna2016super}. 
	 Reconstructions of varying perceptual quality are exemplified with corresponding \ac{PSNR} in \figurename~\ref{fig:exampleIntro}. We illustrate the problem of minimizing \ac{MSE} in \figurename~\ref{fig:manifold} where multiple potential solutions with high texture details are averaged to create a smooth reconstruction.

	In Mathieu et al. \cite{Mathieu2015} and Denton et al. \cite{Denton2015} the authors tackled this problem by employing \acp{GAN} \cite{Goodfellow14GAN} for the application of image generation. Yu and Porikli \cite{yu2016ultra} augment pixel-wise \ac{MSE} loss with a discriminator loss to train a network that super-resolves face images with large upscaling factors ($8\times$). \acp{GAN} were also used for unsupervised representation learning in Radford et al. \cite{Radford2015}.
	The idea of using \acp{GAN} to learn a mapping from one manifold to another is described by Li and Wand \cite{Li2016} for style transfer and Yeh et al. \cite{Yeh2016} for inpainting.
	Bruna et al. \cite{bruna2016super} minimize the squared error in the feature spaces of VGG19 \cite{simonyan2014very} and scattering networks.
	
	Dosovitskiy and Brox \cite{dosovitskiy2016generating} use loss functions based on Euclidean distances computed in the feature space of neural networks in combination with adversarial training. It is shown that the proposed loss allows visually superior image generation and can be used to solve the ill-posed inverse problem of decoding nonlinear feature representations.
	Similar to this work, Johnson et al. \cite{Johnson16PercepLoss} and Bruna et al. \cite{bruna2016super} propose the use of features extracted from a pre-trained VGG network instead of low-level pixel-wise error measures. Specifically the authors formulate a loss function based on the euclidean distance between feature maps extracted from the VGG19 \cite{simonyan2014very} network. Perceptually more convincing results were obtained for both super-resolution and artistic style-transfer \cite{Gatys2015nips,Gatys2016cvpr}. Recently, Li and Wand \cite{Li2016} also investigated the effect of comparing and blending patches in pixel or VGG feature space.

\subsection{Contribution}
	\acp{GAN} provide a powerful framework for generating plausible-looking natural images with high perceptual quality. 
	The \ac{GAN} procedure encourages the reconstructions to move towards regions of the search space with high probability of containing photo-realistic images and thus closer to the natural image manifold as shown in \figurename~\ref{fig:manifold}.

	In this paper we describe the first very deep \ac{ResNet} \cite{he2015deep,he2016identity} architecture using the concept of \acp{GAN} to form a perceptual loss function for photo-realistic \ac{SISR}. Our main contributions are:
	\begin{itemize}
		\item We set a new state of the art for image \ac{SR} with high upscaling factors ($4\times$) as measured by \ac{PSNR} and \ac{SSIM} with our 16 blocks deep \ac{ResNet} (SRResNet) optimized for \ac{MSE}. 
		\item We propose SRGAN which is a \ac{GAN}-based network optimized for a new perceptual loss. Here we replace the \ac{MSE}-based content loss  with a loss calculated on feature maps of the VGG network \cite{simonyan2014very}, which are more invariant to changes in pixel space \cite{Li2016}. 
		\item We confirm with an extensive \ac{MOS} test on images from three public benchmark datasets that SRGAN is the new state of the art, by a large margin, for the estimation of photo-realistic \ac{SR} images with high upscaling factors ($4\times$).
	\end{itemize}


	We describe the network architecture and the perceptual loss in Section \ref{sec:method}. A quantitative evaluation on public benchmark datasets as well as visual illustrations are provided in Section \ref{sec:experiments}. The paper concludes with a discussion in Section \ref{sec:discussion} and concluding remarks in Section \ref{sec:conclusion}.
	
\section{Method}
\label{sec:method}

\begin{figure*}[ht!]
\begin{center}
	  	\begin{tabular}{c}
     	\includegraphics[width=6.5in]{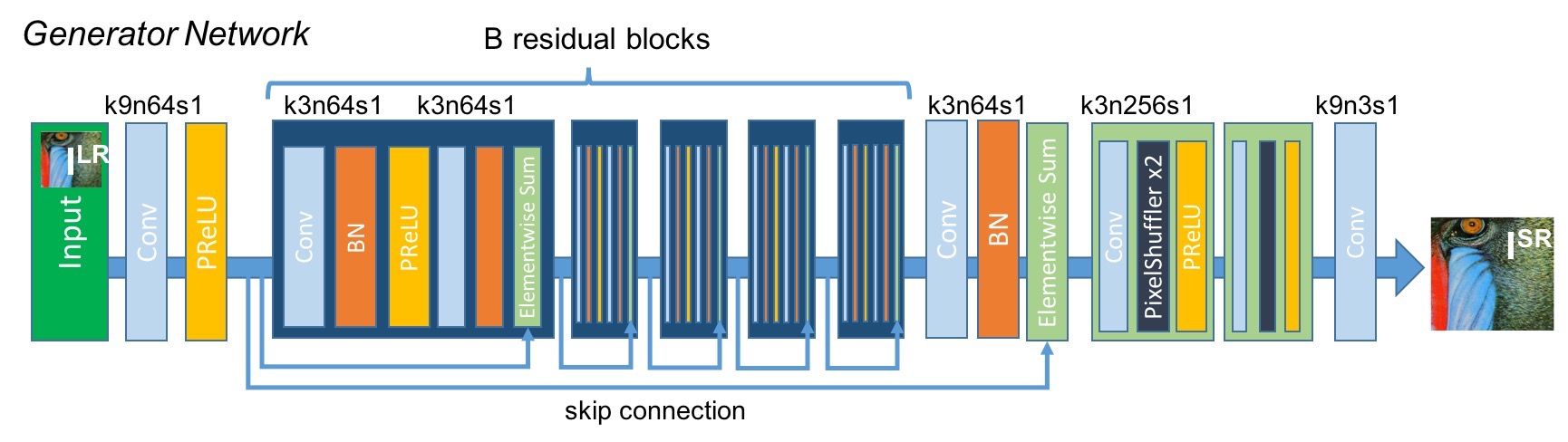}\\
     	\includegraphics[width=6.5in]{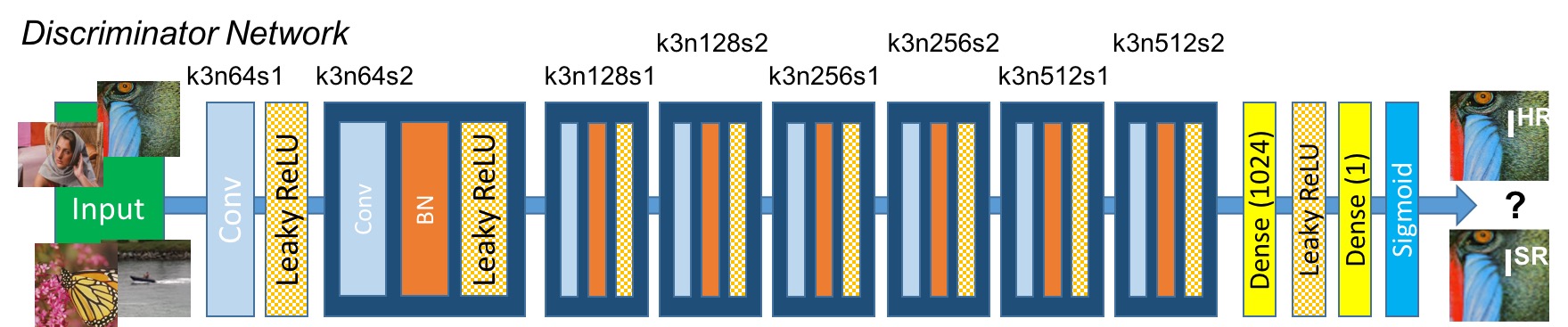}
  	\end{tabular}
\end{center}
  		\caption{Architecture of Generator and Discriminator Network with corresponding kernel size (k), number of feature maps (n) and stride (s) indicated for each convolutional layer.} 
	\label{fig:generator}
\end{figure*}
	In \ac{SISR} the aim is to estimate a high-resolution, super-resolved image $I^{SR}$ from a low-resolution input image $I^{LR}$. Here $I^{LR}$ is the low-resolution version of its high-resolution counterpart $I^{HR}$. The high-resolution images are only available during training. In training, $I^{LR}$ is obtained by applying a Gaussian filter to $I^{HR}$ followed by a downsampling operation with downsampling factor $r$. For an image with $C$ color channels, we describe $I^{LR}$ by a real-valued tensor of size $W \times H \times C$ and $I^{HR}$, $I^{SR}$ by $rW \times rH \times C$ respectively.

	Our ultimate goal is to train a generating function $G$ that estimates for a given \ac{LR} input image its corresponding \ac{HR} counterpart. To achieve this, we train a generator network as a feed-forward \ac{CNN} $G_{\theta_G}$ parametrized by ${\theta_G}$. Here ${\theta_G}=\{W_{1:L}; b_{1:L}\}$ denotes the weights and biases of a $L$-layer deep network and is obtained by optimizing a SR-specific loss function $l^{SR}$. For training images $I^{HR}_n$ , $n = 1,\dots,N$ with corresponding $I^{LR}_n$ , $n = 1,\dots,N$, we solve:
\begin{equation}
	\hat{{\theta}}_G = \arg\min_{\theta_G} \frac{1}{N} \sum_{n=1}^{N}{l^{SR}(G_{\theta_G}(I^{LR}_n),I^{HR}_n)}
\end{equation}
In this work we will specifically design a perceptual loss $l^{SR}$ as a weighted combination of several loss components that model distinct desirable characteristics of the recovered \ac{SR} image. The individual loss functions are described in more detail in Section \ref{sec:losses}.

\subsection{Adversarial network architecture}
Following Goodfellow et al. \cite{Goodfellow14GAN} we further define a discriminator network $D_{\theta_D}$ which we optimize in an alternating manner along with $G_{\theta_G}$ to solve the adversarial min-max problem: 
\begin{equation}
\label{eq:minmax}
\begin{split}
	\min_{\theta_G} \max_{\theta_D} ~& \mathbb{E}_{I^{HR}\sim p_\textrm{train}(I^{HR})} [ \log D_{\theta_D}(I^{HR}) ] + \\
	& \mathbb{E}_{I^{LR}\sim p_G(I^{LR})} [ \log (1-D_{\theta_D}(G_{\theta_G}(I^{LR})) ]
\end{split}
\end{equation}
The general idea behind this formulation is that it allows one to train a generative model $G$ with the goal of fooling a differentiable discriminator $D$ that is trained to distinguish super-resolved images from real images. 
With this approach our generator can learn to create solutions that are highly similar to real images and thus difficult to classify by $D$. This encourages perceptually superior solutions residing in the subspace, the manifold, of natural images. This is in contrast to \ac{SR} solutions obtained by minimizing pixel-wise error measurements, such as the \ac{MSE}.

At the core of our very deep generator network $G$, which is illustrated in \figurename~\ref{fig:generator} are $B$ residual blocks with identical layout. Inspired by Johnson et al. \cite{Johnson16PercepLoss} we employ the block layout proposed by Gross and Wilber \cite{gross2016}. Specifically, we use two convolutional layers with small $3\times3$ kernels and 64 feature maps followed by batch-normalization layers \cite{Ioffe2015} and ParametricReLU \cite{He2015relu} as the activation function.
We increase the resolution of the input image with two trained sub-pixel convolution layers as proposed by Shi et al. \cite{Shi2016ESPCN}.

To discriminate real \ac{HR} images from generated \ac{SR} samples we train a discriminator network. The architecture is shown in \figurename~\ref{fig:generator}.
	We follow the architectural guidelines summarized by Radford et al. \cite{Radford2015} and use LeakyReLU activation ($\alpha=0.2$) and avoid max-pooling throughout the network. The discriminator network is trained to solve the maximization problem in Equation \ref{eq:minmax}. It contains eight convolutional layers with an increasing number of $3\times3$ filter kernels, increasing by a factor of 2 from 64 to 512 kernels as in the VGG network \cite{simonyan2014very}. Strided convolutions are used to reduce the image resolution each time the number of features is doubled. The resulting 512 feature maps are followed by two dense layers and a final sigmoid activation function to obtain a probability for sample classification.

\subsection{Perceptual loss function}
\label{sec:losses}
The definition of our perceptual loss function $l^{SR}$  is critical for the performance of our generator network.
While $l^{SR}$ is commonly modeled based on the \ac{MSE}  \cite{dong2016image,Shi2016ESPCN}, we improve on Johnson et al. \cite{Johnson16PercepLoss} and Bruna et al. \cite{bruna2016super} and design a loss function that assesses a solution with respect to perceptually relevant characteristics.
We formulate the perceptual loss as the weighted sum of a content loss ($l^{SR}_\textrm{X}$) and an adversarial loss component as:

\begin{equation}
	l^{SR} = \underbrace{\underbrace{l^{SR}_\textrm{X}}_{\text{content loss}} + \underbrace{10^{-3} l^{SR}_{Gen}}_{\text{adversarial loss}}}_{\text{perceptual loss (for VGG based content losses)}}
\end{equation}
In the following we describe possible choices for the content loss $l^{SR}_\textrm{X}$ and the adversarial loss $l^{SR}_\textrm{Gen}$.

\subsubsection{Content loss}

The pixel-wise \textbf{\ac{MSE} loss} is calculated as:
\begin{equation}
	l^{SR}_{MSE} = \frac{1}{r^2WH} \sum_{x=1}^{rW} \sum_{y=1}^{rH} (I^{HR}_{x,y} - G_{\theta_G}(I^{LR})_{x,y})^2
\end{equation}
This is the most widely used optimization target for image \ac{SR} on which many state-of-the-art approaches rely \cite{dong2016image,Shi2016ESPCN}. However, while achieving particularly high \ac{PSNR}, solutions of \ac{MSE} optimization problems often lack high-frequency content which results in perceptually unsatisfying solutions with overly smooth textures (\cf \figurename~\ref{fig:exampleIntro}).

Instead of relying on pixel-wise losses we build on the ideas of Gatys et al. \cite{Gatys2015nips}, Bruna et al. \cite{bruna2016super} and Johnson et al. \cite{Johnson16PercepLoss} and use a loss function that is closer to perceptual similarity. We define the \textbf{VGG loss} based on the ReLU activation layers of the pre-trained 19 layer VGG network described in Simonyan and Zisserman \cite{simonyan2014very}. With $\phi_{i,j}$ we indicate the feature map obtained by the j-th convolution (after activation) before the i-th maxpooling layer within the VGG19 network, which we consider given.
	We then define the VGG loss as the euclidean distance between the feature representations of a reconstructed image $G_{\theta_G}(I^{LR})$ and the reference image $I^{HR}$:
	\begin{equation}
	\begin{split}
	l^{SR}_{VGG/i.j} =
	\frac{1}{W_{i,j}H_{i,j}} & \sum_{x=1}^{W_{i,j}} \sum_{y=1}^{H_{i,j}} (\phi_{i,j}(I^{HR})_{x,y} \\
	& - \phi_{i,j}(G_{\theta_G}(I^{LR}))_{x,y})^2	
	\end{split}
	\label{eq:vgg}
	\end{equation}

Here $W_{i,j}$ and $H_{i,j}$ describe the dimensions of the respective feature maps within the VGG network.

\subsubsection{Adversarial loss}
In addition to the content losses described so far, we also add the generative component of our \ac{GAN} to the perceptual loss. This encourages our network to favor solutions that reside on the manifold of natural images, by trying to fool the discriminator network. The generative loss $l^{SR}_{Gen}$ is defined based on the probabilities of the discriminator $D_{\theta_D}(G_{\theta_G}(I^{LR}))$ over all training samples as:
\begin{equation}
	l^{SR}_{Gen} = \sum_{n=1}^{N} -\log D_{\theta_D}(G_{\theta_G}(I^{LR}))
\end{equation}

Here, $D_{\theta_D}(G_{\theta_G}(I^{LR}))$ is the probability that the
reconstructed image $G_{\theta_G}(I^{LR})$ is a natural HR image. For better gradient behavior we minimize $-\log D_{\theta_D}(G_{\theta_G}(I^{LR}))$ instead of $\log [1-D_{\theta_D}(G_{\theta_G}(I^{LR}))]$ \cite{Goodfellow14GAN}.

\section{Experiments}
\label{sec:experiments}
\subsection{Data and similarity measures}
\label{subsec:data}
We perform experiments on three widely used benchmark datasets \textbf{Set5} \cite{bevilacqua2012low}, \textbf{Set14} \cite{zeyde2012single} and \textbf{BSD100}, the testing set of BSD300 \cite{MartinFTM01}. All experiments are performed with a scale factor of $4\times$ between low- and high-resolution images. This corresponds to a $16\times$ reduction in image pixels. For fair  comparison, all reported \ac{PSNR} [dB] and \ac{SSIM} \cite{Wang2004} measures were calculated on the y-channel of center-cropped, removal of a 4-pixel wide strip from each border, images using the daala package\footnote{\url{https://github.com/xiph/daala} (commit: 8d03668)}. Super-resolved images for the reference methods, including nearest neighbor, bicubic, SRCNN \cite{dong2014learning} and SelfExSR \cite{Huang15selfexemplars}, were obtained from online material supplementary to Huang et al.\footnote{\url{https://github.com/jbhuang0604/SelfExSR}} \cite{Huang15selfexemplars} and for DRCN from Kim et al.\footnote{\url{http://cv.snu.ac.kr/research/DRCN/}} \cite{kim2016deeply}.	
	Results obtained with SRResNet (for losses: $l^{SR}_{MSE}$ and $l^{SR}_{VGG/2.2}$) and the SRGAN variants are available online\footnote{\url{https://twitter.box.com/s/lcue6vlrd01ljkdtdkhmfvk7vtjhetog}}.
	Statistical tests were performed as paired two-sided Wilcoxon signed-rank tests and significance determined at $p<0.05$.
	
	The reader may also be interested in an independently developed GAN-based solution on GitHub\footnote{\url{https://github.com/david-gpu/srez}}. However it only provides experimental results on a limited set of faces, which is a more constrained and easier task.

\subsection{Training details and parameters}
We trained all networks on a NVIDIA Tesla M40 GPU using a random sample of 350 thousand images from the \textbf{ImageNet} database \cite{russakovsky2014imagenet}. These images are distinct from the testing images. We obtained the \ac{LR} images by downsampling the \ac{HR} images (BGR, $C=3$) using bicubic kernel with downsampling factor $r=4$. For each mini-batch we crop 16 random $96\times96$ \ac{HR} sub images of distinct training images. Note that we can apply the generator model to images of arbitrary size as it is fully convolutional.
We scaled the range of the \ac{LR} input images to $[0, 1]$ and for the \ac{HR} images to $[-1, 1]$. The \ac{MSE} loss was thus calculated on images of intensity range $[-1, 1]$. VGG feature maps were also rescaled by a factor of $\frac{1}{12.75}$ to obtain VGG losses of a scale that is comparable to the \ac{MSE} loss. This is equivalent to multiplying Equation \ref{eq:vgg} with a rescaling factor of $\approx0.006$.
For optimization we use Adam \cite{Kingma2014} with $\beta_1=0.9$. The SRResNet networks were trained with a learning rate of $10^{-4}$ and $10^{6}$ update iterations. We employed the trained \ac{MSE}-based SRResNet network as initialization for the generator when training the actual \ac{GAN} to avoid undesired local optima.
All SRGAN variants were trained with $10^5$ update iterations at a learning rate of $10^{-4}$ and another $10^5$ iterations at a lower rate of $10^{-5}$.
 	We alternate updates to the generator and discriminator network, which is equivalent to $k=1$ as used in Goodfellow et al. \cite{Goodfellow14GAN}. Our generator network has 16 identical ($B=16$) residual blocks. During test time we turn batch-normalization update off to obtain an output that deterministically depends only on the input \cite{Ioffe2015}. Our implementation is based on Theano \cite{theano2016} and Lasagne \cite{lasagne2015}.

\subsection{Mean opinion score (MOS) testing}
\begin{figure}[ht!] 
     	\includegraphics[width=0.48\textwidth]{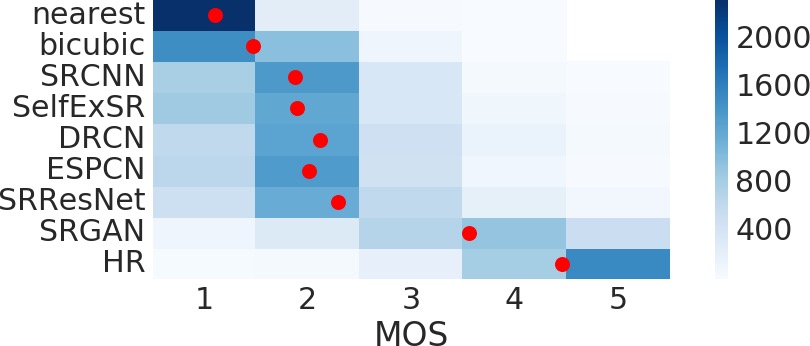}
  		\caption{Color-coded distribution of MOS scores on \textbf{BSD100}. For each method 2600 samples (100 images $\times$ 26 raters) were assessed. Mean shown as red marker, where the bins are centered around value $i$. [$4\times$ upscaling]} 
	\label{fig:MOS}
\end{figure}

We have performed a \ac{MOS} test to quantify the ability of different approaches to reconstruct perceptually convincing images. Specifically, we asked 26 raters to assign an integral score from 1 (bad quality) to 5 (excellent quality) to the super-resolved images. The raters rated 12 versions of each image on \textbf{Set5}, \textbf{Set14} and \textbf{BSD100}: nearest neighbor (NN), bicubic, SRCNN \cite{dong2014learning}, SelfExSR \cite{Huang15selfexemplars}, DRCN \cite{kim2016deeply}, ESPCN \cite{Shi2016ESPCN}, \textbf{SRResNet}-MSE, SRResNet-VGG22$^\ast$ ($^\ast$not rated on \textbf{BSD100}), SRGAN-MSE$^\ast$, SRGAN-VGG22$^\ast$, \textbf{SRGAN}-VGG54 and the original HR image. Each rater thus rated 1128 instances (12 versions of 19 images plus 9 versions of 100 images) that were presented in a randomized fashion.
The raters were calibrated on the NN (score 1) and HR (5) versions of 20 images from the BSD300 training set.
In a pilot study we assessed the calibration procedure and the test-retest reliability of 26 raters on a subset of 10 images from BSD100 by adding a method's images twice to a larger test set. We found good reliability and no significant differences between the ratings of the identical images.
Raters very consistently rated NN interpolated test images as 1 and the original HR images as 5 (\cf \figurename~\ref{fig:MOS}).

The experimental results of the conducted \ac{MOS} tests are summarized in Table \ref{tab:perceptual}, Table \ref{tab:performance} and \figurename~\ref{fig:MOS}.

\begin{table}[]
\centering
\caption{Performance of different loss functions for SRResNet and the adversarial networks on Set5 and Set14 benchmark data. MOS score significantly higher ($p<0.05$) than with other losses in that category$^*$. [$4\times$ upscaling]}
\label{tab:perceptual}
\begin{adjustbox}{max width=0.48\textwidth}
\begin{tabular}{lll | lll}
& \multicolumn{2}{c}{SRResNet-} & \multicolumn{3}{c}{SRGAN-} \\
\textbf{Set5} & MSE & VGG22 & MSE & VGG22 & VGG54 \\
\hline

PSNR &  32.05 & 30.51 & 30.64 & 29.84 & 29.40 \\
SSIM & 0.9019 & 0.8803 & 0.8701 & 0.8468 & 0.8472 \\
MOS  &  3.37 & 3.46 & 3.77 & 3.78 & 3.58 \\ [0.3cm]
\textbf{Set14} & & &  \\
\hline
PSNR &  28.49 & 27.19 & 26.92 & 26.44 & 26.02 \\
SSIM &  0.8184 & 0.7807 & 0.7611 & 0.7518 & 0.7397 \\
MOS  &  2.98 & 3.15$^*$ & 3.43 & 3.57 & 3.72$^*$  \\
\end{tabular}
\end{adjustbox}
\end{table}
\subsection{Investigation of content loss}
\begin{figure*}[ht!]
\begin{tabular}{ cx{3.7cm}cx{3.7cm}cx{3.7cm}cx{3.7cm}cx{3.7cm}}
	 ~~~~~~~~~\textbf{SRResNet} & ~~~~~~~SRGAN-MSE & ~SRGAN-VGG22 & \textbf{SRGAN}-VGG54 & original HR image \\
\end{tabular}

\begin{tikzpicture}[zoomboxarray, zoomboxes below, zoomboxarray inner gap=0.1cm, zoomboxarray columns=2, zoomboxarray rows=2]
    \node [image node] { \includegraphics[trim=200 220 40 5, clip, width=1.3in]{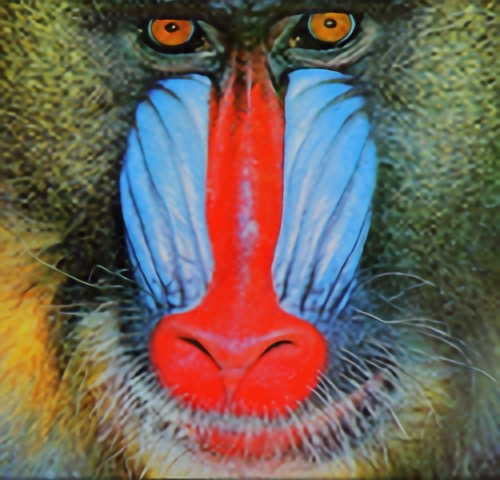} };
    \zoombox[magnification=2,color code=red]{0.5,0.85}
    \zoombox[magnification=2,color code=yellow]{0.7,0.55}
    \zoombox[magnification=2,color code=orange]{0.35,0.3}
    \zoombox[magnification=5,color code=lime]{0.86,0.25}
\end{tikzpicture}
\begin{tikzpicture}[zoomboxarray, zoomboxes below, zoomboxarray inner gap=0.1cm, zoomboxarray columns=2, zoomboxarray rows=2]
    \node [image node] { \includegraphics[trim=200 220 40 5, clip, width=1.3in]{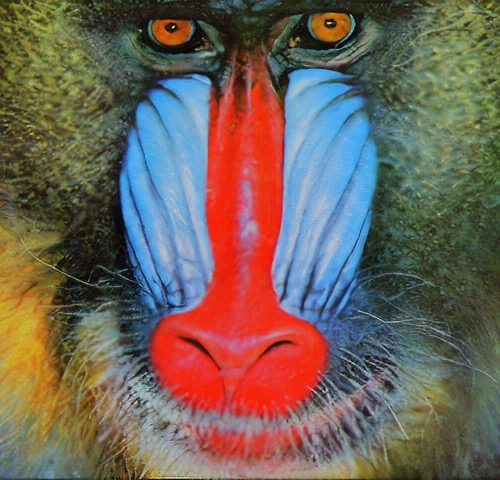} };
    \zoombox[magnification=2,color code=red]{0.5,0.85}
    \zoombox[magnification=2,color code=yellow]{0.7,0.55}
    \zoombox[magnification=2,color code=orange]{0.35,0.3}
    \zoombox[magnification=5,color code=lime]{0.86,0.25}
\end{tikzpicture}
\begin{tikzpicture}[zoomboxarray, zoomboxes below, zoomboxarray inner gap=0.1cm, zoomboxarray columns=2, zoomboxarray rows=2]
    \node [image node] { \includegraphics[trim=200 220 40 5, clip, width=1.3in]{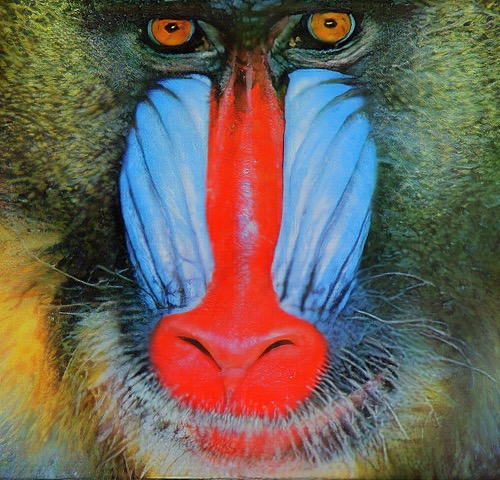} };
    \zoombox[magnification=2,color code=red]{0.5,0.85}
    \zoombox[magnification=2,color code=yellow]{0.7,0.55}
    \zoombox[magnification=2,color code=orange]{0.35,0.3}
    \zoombox[magnification=5,color code=lime]{0.86,0.25}
\end{tikzpicture}
\begin{tikzpicture}[zoomboxarray, zoomboxes below, zoomboxarray inner gap=0.1cm, zoomboxarray columns=2, zoomboxarray rows=2]
    \node [image node] { \includegraphics[trim=200 220 40 5, clip, width=1.3in]{images/used/jpg/baboon_SRGAN-VGG54} };
    \zoombox[magnification=2,color code=red]{0.5,0.85}
    \zoombox[magnification=2,color code=yellow]{0.7,0.55}
    \zoombox[magnification=2,color code=orange]{0.35,0.3}
    \zoombox[magnification=5,color code=lime]{0.86,0.25}
\end{tikzpicture}
\begin{tikzpicture}[zoomboxarray, zoomboxes below, zoomboxarray inner gap=0.1cm, zoomboxarray columns=2, zoomboxarray rows=2]
    \node [image node] { \includegraphics[trim=200 220 40 5, clip, width=1.3in]{images/used/jpg/baboon_HR} };
    \zoombox[magnification=2,color code=red]{0.5,0.85}
    \zoombox[magnification=2,color code=yellow]{0.7,0.55}
    \zoombox[magnification=2,color code=orange]{0.35,0.3}
    \zoombox[magnification=5,color code=lime]{0.86,0.25}
\end{tikzpicture}
\caption{\textbf{SRResNet} (left: a,b), SRGAN-MSE (middle left: c,d), SRGAN-VGG2.2 (middle: e,f) and  \textbf{SRGAN}-VGG54 (middle right: g,h) reconstruction results and corresponding reference HR image (right: i,j). [$4\times$ upscaling]}
\label{fig:perceptual}
\end{figure*}

We investigated the effect of different content loss choices in the perceptual loss for the \ac{GAN}-based networks. Specifically we investigate $l^{SR} = l^{SR}_\textrm{X} + 10^{-3}l^{SR}_{Gen}$ for the following content losses $l^{SR}_\textrm{X}$:\\
\begin{itemize}
	\item SRGAN-MSE: $l^{SR}_{MSE}$, to investigate the adversarial network with the standard \ac{MSE} as content loss.
 	\item SRGAN-VGG22: $l^{SR}_{VGG/2.2}$ with $\phi_{2,2}$, a loss defined on feature maps representing lower-level features \cite{zeiler2014visualizing}.
	\item \textbf{SRGAN}-VGG54: $l^{SR}_{VGG/5.4}$ with $\phi_{5,4}$, a loss defined on feature maps of higher level features from deeper network layers with more potential to focus on the content of the images \cite{zeiler2014visualizing,Yosinski2015,Mahendran2016}. We refer to this network as \textbf{SRGAN} in the following.
\end{itemize}
We also evaluate the performance of the generator network without adversarial component for the two losses $l^{SR}_{MSE}$ (\textbf{SRResNet}-MSE) and $l^{SR}_{VGG/2.2}$ (SRResNet-VGG22). We refer to SRResNet-MSE as \textbf{SRResNet}.
Note, when training SRResNet-VGG22 we added an additional total variation loss with weight $2\times10^{-8}$ to $l^{SR}_{VGG/2.2}$ \cite{Aly2005,Johnson16PercepLoss}. 
 Quantitative results are summarized in Table \ref{tab:perceptual} and visual examples provided in \figurename~\ref{fig:perceptual}.
 Even combined with the adversarial loss, \ac{MSE} provides solutions with the highest \ac{PSNR} values that are, however, perceptually rather smooth and less convincing than results achieved with a loss component more sensitive to visual perception. This is caused by competition between the \ac{MSE}-based content loss and the adversarial loss. We further attribute minor reconstruction artifacts, which we observed in a minority of SRGAN-MSE-based reconstructions, to those competing objectives. We could not determine a significantly best loss function for SRResNet or SRGAN with respect to \ac{MOS} score on \textbf{Set5}. However, \textbf{SRGAN}-VGG54 significantly outperformed other SRGAN and SRResNet variants on \textbf{Set14} in terms of \ac{MOS}.
 	We observed a trend that using the higher level VGG feature maps $\phi_{5,4}$ yields better texture detail when compared to $\phi_{2,2}$ (\cf \figurename~\ref{fig:perceptual}).
	Further examples of perceptual improvements through \textbf{SRGAN} over \textbf{SRResNet} are provided in the supplementary material.

\begin{table*}[htb!]
\centering
\caption{Comparison of NN, bicubic, SRCNN \cite{dong2014learning}, SelfExSR \cite{Huang15selfexemplars}, DRCN \cite{kim2016deeply}, ESPCN \cite{Shi2016ESPCN}, \textbf{SRResNet}, \textbf{SRGAN}-VGG54 and the original HR on benchmark data. Highest measures (PSNR [dB], SSIM, MOS) in bold. [$4\times$ upscaling]}
\label{tab:performance}
\begin{adjustbox}{max width=0.8\textwidth}
\begin{tabular}{llllllllll}
\textbf{Set5} & nearest & bicubic & SRCNN & SelfExSR & DRCN & ESPCN & \textbf{SRResNet} & \textbf{SRGAN} & HR \\
\hline
PSNR & 26.26 & 28.43 & 30.07 & 30.33 & 31.52 & 30.76 & \textbf{32.05} & 29.40 & $\infty$ \\
SSIM & 0.7552 & 0.8211 & 0.8627 & 0.872 & 0.8938 & 0.8784 & \textbf{0.9019} & 0.8472 & 1 \\
MOS  & 1.28 & 1.97 & 2.57 & 2.65 & 3.26 & 2.89 & 3.37 & \textbf{3.58} & 4.32 \\ [0.3cm]

\textbf{Set14} & & & & & & & & \\
\hline
PSNR & 24.64 & 25.99 & 27.18 & 27.45 & 28.02 & 27.66 & \textbf{28.49} & 26.02 & $\infty$ \\
SSIM & 0.7100 & 0.7486 & 0.7861 & 0.7972 & 0.8074 & 0.8004 & \textbf{0.8184} & 0.7397 & 1 \\
MOS  & 1.20 & 1.80 & 2.26 & 2.34 & 2.84 & 2.52 & 2.98 & \textbf{3.72} & 4.32 \\ [0.3cm]
\textbf{BSD100} & & & & & & & &  \\
\hline
PSNR & 25.02 & 25.94 & 26.68 & 26.83 & 27.21 & 27.02 & \textbf{27.58} & 25.16 & $\infty$ \\
SSIM & 0.6606 & 0.6935 & 0.7291 & 0.7387 & 0.7493 & 0.7442 & \textbf{0.7620} & 0.6688 & 1 \\
MOS  & 1.11 & 1.47 & 1.87 & 1.89 & 2.12 & 2.01 & 2.29 & \textbf{3.56} & 4.46 \\
\end{tabular}
\end{adjustbox}
\end{table*}

\subsection{Performance of the final networks}
\label{subsec:msebased}
We compare the performance of \textbf{SRResNet} and \textbf{SRGAN} to NN, bicubic interpolation, and four state-of-the-art methods.
Quantitative results are summarized in Table \ref{tab:performance} and confirm that \textbf{SRResNet} (in terms of PSNR/SSIM) sets a new state of the art on three benchmark datasets.
Please note that we used a publicly available framework for evaluation (\cf Section \ref{subsec:data}), reported values might thus slightly deviate from those reported in the original papers.

We further obtained \ac{MOS} ratings for \textbf{SRGAN} and all reference methods on \textbf{BSD100}. Examples of images super-resolved with \textbf{SRResNet} and \textbf{SRGAN} are depicted in the supplementary material.
The results shown in Table \ref{tab:performance} confirm that \textbf{SRGAN} outperforms all reference methods by a large margin and sets a new state of the art for photo-realistic image \ac{SR}. All differences in \ac{MOS} (\cf Table \ref{tab:performance}) are highly significant on \textbf{BSD100}, except SRCNN vs. SelfExSR. The distribution of all collected \ac{MOS} ratings is summarized in \figurename~\ref{fig:MOS}.

\section{Discussion and future work}
\label{sec:discussion}
We confirmed the superior perceptual performance of \textbf{SRGAN} using \ac{MOS} testing. We have further shown that standard quantitative measures such as PSNR and SSIM fail to capture and accurately assess image quality with respect to the human visual system \cite{Toderici2016}. 
The focus of this work was the perceptual quality of super-resolved images rather than computational efficiency. The presented model is, in contrast to Shi et al. \cite{Shi2016ESPCN}, not optimized for video \ac{SR} in real-time. However, preliminary experiments on the network architecture suggest that shallower networks have the potential to provide very efficient alternatives at a small reduction of qualitative performance. 
In contrast to Dong et al. \cite{dong2016image}, we found deeper network architectures to be beneficial. We speculate that the \ac{ResNet} design has a substantial impact on the performance of deeper networks. We found that even deeper  networks ($B>16$) can further increase the performance of \textbf{SRResNet}, however, come at the cost of longer training and testing times (\cf supplementary material). We further found \ac{SRGAN} variants of deeper networks are increasingly difficult to train due to the appearance of high-frequency artifacts. 

Of particular importance when aiming for photo-realistic solutions to the \ac{SR} problem is the choice of the content loss as illustrated in \figurename~\ref{fig:perceptual}. In this work, we found $l^{SR}_{VGG/5.4}$ to yield the perceptually most convincing results, which we attribute to the potential of deeper network layers to represent features of higher abstraction \cite{zeiler2014visualizing,Yosinski2015,Mahendran2016} away from pixel space. We speculate that feature maps of these deeper layers focus purely on the content while leaving the adversarial loss focusing on texture details which are the main difference between the super-resolved images without the adversarial loss and photo-realistic images.
We also note that the ideal loss function depends on the application. For example, approaches that hallucinate finer detail might be less suited for medical applications or surveillance. The perceptually convincing reconstruction of text or structured scenes \cite{Huang15selfexemplars} is challenging and part of future work.
The development of content loss functions that describe image spatial content, but more invariant to changes in pixel space will further improve photo-realistic image \ac{SR} results.
\section{Conclusion}
\label{sec:conclusion}
We have described a deep residual network \textbf{SRResNet} that sets a new state of the art on public benchmark datasets when evaluated with the widely used \ac{PSNR} measure. We have highlighted some limitations of this \ac{PSNR}-focused image super-resolution and introduced \textbf{SRGAN}, which augments the content loss function with an adversarial loss by training a \ac{GAN}. Using extensive \ac{MOS} testing, we have confirmed that \textbf{SRGAN} reconstructions for large upscaling factors ($4\times$) are, by a considerable margin, more photo-realistic than reconstructions obtained with state-of-the-art reference methods. 

{\footnotesize
\bibliographystyle{ieee}
\bibliography{bibliograph}

\begin{thebibliography}{10}\itemsep=-1pt

\bibitem{Allebach96}
J.~Allebach and P.~W. Wong.
\newblock Edge-directed interpolation.
\newblock In {\em Proceedings of International Conference on Image Processing},
  volume~3, pages 707--710, 1996.

\bibitem{Aly2005}
H.~A. Aly and E.~Dubois.
\newblock {Image up-sampling using total-variation regularization with a new
  observation model}.
\newblock {\em IEEE Transactions on Image Processing}, 14(10):1647--1659, 2005.

\bibitem{bevilacqua2012low}
M.~Bevilacqua, A.~Roumy, C.~Guillemot, and M.~L. Alberi-Morel.
\newblock Low-complexity single-image super-resolution based on nonnegative
  neighbor embedding.
\newblock {\em BMVC}, 2012.

\bibitem{Borman1998aSurvey}
S.~Borman and R.~L. Stevenson.
\newblock {Super-Resolution from Image Sequences - A Review}.
\newblock {\em Midwest Symposium on Circuits and Systems}, pages 374--378,
  1998.

\bibitem{bruna2016super}
J.~Bruna, P.~Sprechmann, and Y.~LeCun.
\newblock Super-resolution with deep convolutional sufficient statistics.
\newblock In {\em International Conference on Learning Representations (ICLR)},
  2016.

\bibitem{dai2015jointly}
D.~Dai, R.~Timofte, and L.~Van~Gool.
\newblock Jointly optimized regressors for image super-resolution.
\newblock In {\em Computer Graphics Forum}, volume~34, pages 95--104, 2015.

\bibitem{Denton2015}
E.~Denton, S.~Chintala, A.~Szlam, and R.~Fergus.
\newblock Deep generative image models using a laplacian pyramid of adversarial
  networks.
\newblock In {\em Advances in Neural Information Processing Systems (NIPS)},
  pages 1486--1494, 2015.

\bibitem{lasagne2015}
S.~Dieleman, J.~Schl{\"{u}}ter, C.~Raffel, E.~Olson, S.~K. Sønderby, D.~Nouri,
  D.~Maturana, M.~Thoma, E.~Battenberg, J.~Kelly, J.~D. Fauw, M.~Heilman,
  diogo149, B.~McFee, H.~Weideman, takacsg84, peterderivaz, Jon, instagibbs,
  D.~K. Rasul, CongLiu, Britefury, and J.~Degrave.
\newblock Lasagne: First release., 2015.

\bibitem{dong2014learning}
C.~Dong, C.~C. Loy, K.~He, and X.~Tang.
\newblock Learning a deep convolutional network for image super-resolution.
\newblock In {\em European Conference on Computer Vision (ECCV)}, pages
  184--199. Springer, 2014.

\bibitem{dong2016image}
C.~Dong, C.~C. Loy, K.~He, and X.~Tang.
\newblock Image super-resolution using deep convolutional networks.
\newblock {\em IEEE Transactions on Pattern Analysis and Machine Intelligence},
  38(2):295--307, 2016.

\bibitem{dong2016accelerating}
C.~Dong, C.~C. Loy, and X.~Tang.
\newblock Accelerating the super-resolution convolutional neural network.
\newblock In {\em European Conference on Computer Vision (ECCV)}, pages
  391--407. Springer, 2016.

\bibitem{Dong2011}
W.~Dong, L.~Zhang, G.~Shi, and X.~Wu.
\newblock {Image deblurring and super-resolution by adaptive sparse domain
  selection and adaptive regularization}.
\newblock {\em IEEE Transactions on Image Processing}, 20(7):1838--1857, 2011.

\bibitem{dosovitskiy2016generating}
A.~Dosovitskiy and T.~Brox.
\newblock Generating images with perceptual similarity metrics based on deep
  networks.
\newblock In {\em Advances in Neural Information Processing Systems (NIPS)},
  pages 658--666, 2016.

\bibitem{Duchon1979}
C.~E. Duchon.
\newblock {Lanczos Filtering in One and Two Dimensions}.
\newblock In {\em Journal of Applied Meteorology}, volume~18, pages 1016--1022.
  1979.

\bibitem{Farsiu2004}
S.~Farsiu, M.~D. Robinson, M.~Elad, and P.~Milanfar.
\newblock {Fast and robust multiframe super resolution}.
\newblock {\em IEEE Transactions on Image Processing}, 13(10):1327--1344, 2004.

\bibitem{Ferwerda2003}
J.~A. Ferwerda.
\newblock Three varieties of realism in computer graphics.
\newblock In {\em Electronic Imaging}, pages 290--297. International Society
  for Optics and Photonics, 2003.

\bibitem{Freeman2002}
W.~T. Freeman, T.~R. Jones, and E.~C. Pasztor.
\newblock {Example-based super-resolution}.
\newblock {\em IEEE Computer Graphics and Applications}, 22(2):56--65, 2002.

\bibitem{Freeman2000}
W.~T. Freeman, E.~C. Pasztor, and O.~T. Carmichael.
\newblock Learning low-level vision.
\newblock {\em International Journal of Computer Vision}, 40(1):25--47, 2000.

\bibitem{Gatys2015nips}
L.~A. Gatys, A.~S. Ecker, and M.~Bethge.
\newblock Texture synthesis using convolutional neural networks.
\newblock In {\em Advances in Neural Information Processing Systems (NIPS)},
  pages 262--270, 2015.

\bibitem{Gatys2016cvpr}
L.~A. Gatys, A.~S. Ecker, and M.~Bethge.
\newblock {Image Style Transfer Using Convolutional Neural Networks}.
\newblock In {\em IEEE Conference on Computer Vision and Pattern Recognition
  (CVPR)}, pages 2414--2423, 2016.

\bibitem{glasner2009super}
D.~Glasner, S.~Bagon, and M.~Irani.
\newblock Super-resolution from a single image.
\newblock In {\em IEEE International Conference on Computer Vision (ICCV)},
  pages 349--356, 2009.

\bibitem{Goodfellow14GAN}
I.~Goodfellow, J.~Pouget-Abadie, M.~Mirza, B.~Xu, D.~Warde-Farley, S.~Ozair,
  A.~Courville, and Y.~Bengio.
\newblock Generative adversarial nets.
\newblock In {\em Advances in Neural Information Processing Systems (NIPS)},
  pages 2672--2680, 2014.

\bibitem{gregor2010learning}
K.~Gregor and Y.~LeCun.
\newblock Learning fast approximations of sparse coding.
\newblock In {\em Proceedings of the 27th International Conference on Machine
  Learning (ICML-10)}, pages 399--406, 2010.

\bibitem{gross2016}
S.~Gross and M.~Wilber.
\newblock Training and investigating residual nets, online at
  \url{http://torch.ch/blog/2016/02/04/resnets.html}.
\newblock 2016.

\bibitem{gu2015convolutional}
S.~Gu, W.~Zuo, Q.~Xie, D.~Meng, X.~Feng, and L.~Zhang.
\newblock Convolutional sparse coding for image super-resolution.
\newblock In {\em IEEE International Conference on Computer Vision (ICCV)},
  pages 1823--1831. 2015.

\bibitem{Gupta2011}
P.~Gupta, P.~Srivastava, S.~Bhardwaj, and V.~Bhateja.
\newblock A modified psnr metric based on hvs for quality assessment of color
  images.
\newblock In {\em IEEE International Conference on Communication and Industrial
  Application (ICCIA)}, pages 1--4, 2011.

\bibitem{he2011single}
H.~He and W.-C. Siu.
\newblock Single image super-resolution using gaussian process regression.
\newblock In {\em IEEE Conference on Computer Vision and Pattern Recognition
  (CVPR)}, pages 449--456, 2011.

\bibitem{He2015relu}
K.~He, X.~Zhang, S.~Ren, and J.~Sun.
\newblock Delving deep into rectifiers: Surpassing human-level performance on
  imagenet classification.
\newblock In {\em IEEE International Conference on Computer Vision (ICCV)},
  pages 1026--1034, 2015.

\bibitem{he2015deep}
K.~He, X.~Zhang, S.~Ren, and J.~Sun.
\newblock Deep residual learning for image recognition.
\newblock In {\em IEEE Conference on Computer Vision and Pattern Recognition
  (CVPR)}, pages 770--778, 2016.

\bibitem{he2016identity}
K.~He, X.~Zhang, S.~Ren, and J.~Sun.
\newblock Identity mappings in deep residual networks.
\newblock In {\em European Conference on Computer Vision (ECCV)}, pages
  630--645. Springer, 2016.

\bibitem{Huang15selfexemplars}
J.~B. Huang, A.~Singh, and N.~Ahuja.
\newblock Single image super-resolution from transformed self-exemplars.
\newblock In {\em IEEE Conference on Computer Vision and Pattern Recognition
  (CVPR)}, pages 5197--5206, 2015.

\bibitem{Ioffe2015}
S.~Ioffe and C.~Szegedy.
\newblock Batch normalization: Accelerating deep network training by reducing
  internal covariate shift.
\newblock In {\em Proceedings of The 32nd International Conference on Machine
  Learning (ICML)}, pages 448--456, 2015.

\bibitem{Johnson16PercepLoss}
J.~Johnson, A.~Alahi, and F.~Li.
\newblock Perceptual losses for real-time style transfer and super- resolution.
\newblock In {\em European Conference on Computer Vision (ECCV)}, pages
  694--711. Springer, 2016.

\bibitem{kim2016deeply}
J.~Kim, J.~K. Lee, and K.~M. Lee.
\newblock Deeply-recursive convolutional network for image super-resolution.
\newblock In {\em IEEE Conference on Computer Vision and Pattern Recognition
  (CVPR)}, 2016.

\bibitem{Kim10kernelregression}
K.~I. Kim and Y.~Kwon.
\newblock Single-image super-resolution using sparse regression and natural
  image prior.
\newblock {\em IEEE Transactions on Pattern Analysis and Machine Intelligence},
  32(6):1127--1133, 2010.

\bibitem{Kingma2014}
D.~Kingma and J.~Ba.
\newblock Adam: A method for stochastic optimization.
\newblock In {\em International Conference on Learning Representations (ICLR)},
  2015.

\bibitem{krizhevsky2012imagenet}
A.~Krizhevsky, I.~Sutskever, and G.~E. Hinton.
\newblock Imagenet classification with deep convolutional neural networks.
\newblock In {\em Advances in Neural Information Processing Systems (NIPS)},
  pages 1097--1105, 2012.

\bibitem{Li2016}
C.~Li and M.~Wand.
\newblock {Combining Markov Random Fields and Convolutional Neural Networks for
  Image Synthesis}.
\newblock In {\em IEEE Conference on Computer Vision and Pattern Recognition
  (CVPR)}, pages 2479--2486, 2016.

\bibitem{Li2001}
X.~Li and M.~T. Orchard.
\newblock {New edge-directed interpolation}.
\newblock {\em IEEE Transactions on Image Processing}, 10(10):1521--1527, 2001.

\bibitem{Mahendran2016}
A.~Mahendran and A.~Vedaldi.
\newblock Visualizing deep convolutional neural networks using natural
  pre-images.
\newblock {\em International Journal of Computer Vision}, pages 1--23, 2016.

\bibitem{MartinFTM01}
D.~Martin, C.~Fowlkes, D.~Tal, and J.~Malik.
\newblock A database of human segmented natural images and its application to
  evaluating segmentation algorithms and measuring ecological statistics.
\newblock In {\em IEEE International Conference on Computer Vision (ICCV)},
  volume~2, pages 416--423, 2001.

\bibitem{Mathieu2015}
M.~Mathieu, C.~Couprie, and Y.~LeCun.
\newblock Deep multi-scale video prediction beyond mean square error.
\newblock In {\em International Conference on Learning Representations (ICLR)},
  2016.

\bibitem{Nasrollahi2014}
K.~Nasrollahi and T.~B. Moeslund.
\newblock {Super-resolution: A comprehensive survey}.
\newblock In {\em Machine Vision and Applications}, volume~25, pages
  1423--1468. 2014.

\bibitem{Radford2015}
A.~Radford, L.~Metz, and S.~Chintala.
\newblock Unsupervised representation learning with deep convolutional
  generative adversarial networks.
\newblock In {\em International Conference on Learning Representations (ICLR)},
  2016.

\bibitem{russakovsky2014imagenet}
O.~Russakovsky, J.~Deng, H.~Su, J.~Krause, S.~Satheesh, S.~Ma, Z.~Huang,
  A.~Karpathy, A.~Khosla, M.~Bernstein, et~al.
\newblock Imagenet large scale visual recognition challenge.
\newblock {\em International Journal of Computer Vision}, pages 1--42, 2014.

\bibitem{salvador2015naive}
J.~Salvador and E.~P{\'e}rez-Pellitero.
\newblock Naive bayes super-resolution forest.
\newblock In {\em IEEE International Conference on Computer Vision (ICCV)},
  pages 325--333. 2015.

\bibitem{schulter2015fast}
S.~Schulter, C.~Leistner, and H.~Bischof.
\newblock Fast and accurate image upscaling with super-resolution forests.
\newblock In {\em IEEE Conference on Computer Vision and Pattern Recognition
  (CVPR)}, pages 3791--3799, 2015.

\bibitem{Shi2016ESPCN}
W.~Shi, J.~Caballero, F.~Huszar, J.~Totz, A.~P. Aitken, R.~Bishop, D.~Rueckert,
  and Z.~Wang.
\newblock {Real-Time Single Image and Video Super-Resolution Using an Efficient
  Sub-Pixel Convolutional Neural Network}.
\newblock In {\em IEEE Conference on Computer Vision and Pattern Recognition
  (CVPR)}, pages 1874--1883, 2016.

\bibitem{simonyan2014very}
K.~Simonyan and A.~Zisserman.
\newblock Very deep convolutional networks for large-scale image recognition.
\newblock In {\em International Conference on Learning Representations (ICLR)},
  2015.

\bibitem{Sun2008}
J.~Sun, J.~Sun, Z.~Xu, and H.-Y. Shum.
\newblock Image super-resolution using gradient profile prior.
\newblock In {\em IEEE Conference on Computer Vision and Pattern Recognition
  (CVPR)}, pages 1--8, 2008.

\bibitem{szegedy2015going}
C.~Szegedy, W.~Liu, Y.~Jia, P.~Sermanet, S.~Reed, D.~Anguelov, D.~Erhan,
  V.~Vanhoucke, and A.~Rabinovich.
\newblock Going deeper with convolutions.
\newblock In {\em IEEE Conference on Computer Vision and Pattern Recognition
  (CVPR)}, pages 1--9, 2015.

\bibitem{Tai2010}
Y.-W. Tai, S.~Liu, M.~S. Brown, and S.~Lin.
\newblock {Super Resolution using Edge Prior and Single Image Detail
  Synthesis}.
\newblock In {\em IEEE Conference on Computer Vision and Pattern Recognition
  (CVPR)}, pages 2400--2407, 2010.

\bibitem{theano2016}
{Theano Development Team}.
\newblock {Theano: A {Python} framework for fast computation of mathematical
  expressions}.
\newblock {\em arXiv preprint arXiv:1605.02688}, 2016.

\bibitem{timofte2013anchored}
R.~Timofte, V.~De, and L.~Van~Gool.
\newblock Anchored neighborhood regression for fast example-based
  super-resolution.
\newblock In {\em IEEE International Conference on Computer Vision (ICCV)},
  pages 1920--1927, 2013.

\bibitem{timofte2014a+}
R.~Timofte, V.~De~Smet, and L.~Van~Gool.
\newblock A+: Adjusted anchored neighborhood regression for fast
  super-resolution.
\newblock In {\em Asian Conference on Computer Vision (ACCV)}, pages 111--126.
  Springer, 2014.

\bibitem{Toderici2016}
G.~Toderici, D.~Vincent, N.~Johnston, S.~J. Hwang, D.~Minnen, J.~Shor, and
  M.~Covell.
\newblock {Full Resolution Image Compression with Recurrent Neural Networks}.
\newblock {\em arXiv preprint arXiv:1608.05148}, 2016.

\bibitem{Wang2016}
Y.~Wang, L.~Wang, H.~Wang, and P.~Li.
\newblock {End-to-End Image Super-Resolution via Deep and Shallow Convolutional
  Networks}.
\newblock {\em arXiv preprint arXiv:1607.07680}, 2016.

\bibitem{Wang2004}
Z.~Wang, A.~C. Bovik, H.~R. Sheikh, and E.~P. Simoncelli.
\newblock {Image quality assessment: From error visibility to structural
  similarity}.
\newblock {\em IEEE Transactions on Image Processing}, 13(4):600--612, 2004.

\bibitem{Wang2015}
Z.~Wang, D.~Liu, J.~Yang, W.~Han, and T.~Huang.
\newblock Deep networks for image super-resolution with sparse prior.
\newblock In {\em IEEE International Conference on Computer Vision (ICCV)},
  pages 370--378, 2015.

\bibitem{Wang2003}
Z.~Wang, E.~P. Simoncelli, and A.~C. Bovik.
\newblock {Multi-scale structural similarity for image quality assessment}.
\newblock In {\em IEEE Asilomar Conference on Signals, Systems and Computers},
  volume~2, pages 9--13, 2003.

\bibitem{Yang14benchmark}
C.-Y. Yang, C.~Ma, and M.-H. Yang.
\newblock Single-image super-resolution: A benchmark.
\newblock In {\em European Conference on Computer Vision (ECCV)}, pages
  372--386. Springer, 2014.

\bibitem{Yang08}
J.~Yang, J.~Wright, T.~Huang, and Y.~Ma.
\newblock Image super-resolution as sparse representation of raw image patches.
\newblock In {\em IEEE Conference on Computer Vision and Pattern Recognition
  (CVPR)}, pages 1--8, 2008.

\bibitem{yang2007spatial}
Q.~Yang, R.~Yang, J.~Davis, and D.~Nist{\'e}r.
\newblock Spatial-depth super resolution for range images.
\newblock In {\em IEEE Conference on Computer Vision and Pattern Recognition
  (CVPR)}, pages 1--8, 2007.

\bibitem{Yeh2016}
R.~Yeh, C.~Chen, T.~Y. Lim, M.~Hasegawa-Johnson, and M.~N. Do.
\newblock {Semantic Image Inpainting with Perceptual and Contextual Losses}.
\newblock {\em arXiv preprint arXiv:1607.07539}, 2016.

\bibitem{Yosinski2015}
J.~Yosinski, J.~Clune, A.~Nguyen, T.~Fuchs, and H.~Lipson.
\newblock {Understanding Neural Networks Through Deep Visualization}.
\newblock In {\em International Conference on Machine Learning - Deep Learning
  Workshop 2015}, page~12, 2015.

\bibitem{yu2016ultra}
X.~Yu and F.~Porikli.
\newblock Ultra-resolving face images by discriminative generative networks.
\newblock In {\em European Conference on Computer Vision (ECCV)}, pages
  318--333. 2016.

\bibitem{Yue2013}
H.~Yue, X.~Sun, J.~Yang, and F.~Wu.
\newblock {Landmark image super-resolution by retrieving web images}.
\newblock {\em IEEE Transactions on Image Processing}, 22(12):4865--4878, 2013.

\bibitem{zeiler2014visualizing}
M.~D. Zeiler and R.~Fergus.
\newblock Visualizing and understanding convolutional networks.
\newblock In {\em European Conference on Computer Vision (ECCV)}, pages
  818--833. Springer, 2014.

\bibitem{zeyde2012single}
R.~Zeyde, M.~Elad, and M.~Protter.
\newblock On single image scale-up using sparse-representations.
\newblock In {\em Curves and Surfaces}, pages 711--730. Springer, 2012.

\bibitem{zhang2012multi}
K.~Zhang, X.~Gao, D.~Tao, and X.~Li.
\newblock Multi-scale dictionary for single image super-resolution.
\newblock In {\em IEEE Conference on Computer Vision and Pattern Recognition
  (CVPR)}, pages 1114--1121, 2012.

\bibitem{Zou12}
W.~Zou and P.~C. Yuen.
\newblock {Very Low Resolution Face Recognition in Parallel Environment }.
\newblock {\em IEEE Transactions on Image Processing}, 21:327--340, 2012.

\end{thebibliography}
}
\clearpage
\setcounter{section}{0}
\renewcommand{\thesection}{\Alph{section}}
\onecolumn
\section{Supplementary Material}
In this supplementary material we first briefly investigate the influence of network depth (number of residual blocks) on the performance (PSNR, time) of \textbf{SRResNet} in Section \ref{app:performance}. We then visualize on an example image how the \textbf{SRGAN} network performance evolves with increasing number of training iterations in Section \ref{app:evolution}. Results of the \ac{MOS} tests conducted on \textbf{Set5}, \textbf{Set14}, \textbf{BSD100} are summarized in Section \ref{app:MOS}. Finally we provide a visualization of all image reconstruction obtained with \textbf{SRResNet} and \textbf{SRGAN} with a $4\times$ upscaling factor for \textbf{Set5} (Section \ref{app:Set5}), \textbf{Set14} (Section \ref{app:Set14}) and five randomly selected images from \textbf{BSD100} (Section \ref{app:BSD100}).\\

Images are best viewed and compared zoomed in. All original low-/high-resolution images and reconstructions ($4\times$ upscaling) obtained with different methods (bicubic, SRResNet-MSE, SRResNet-VGG22, SRGAN-MSE, SRGAN-VGG22, SRGAN-VGG54) described in the paper are available for download at \url{https://twitter.box.com/s/lcue6vlrd01ljkdtdkhmfvk7vtjhetog}.

\clearpage

\subsection{Performance (PSNR/time) vs. network depth}
\label{app:performance}
\begin{figure*}[ht!] 
  	\begin{tabular}{cc}
     	\includegraphics[width=0.5\textwidth]{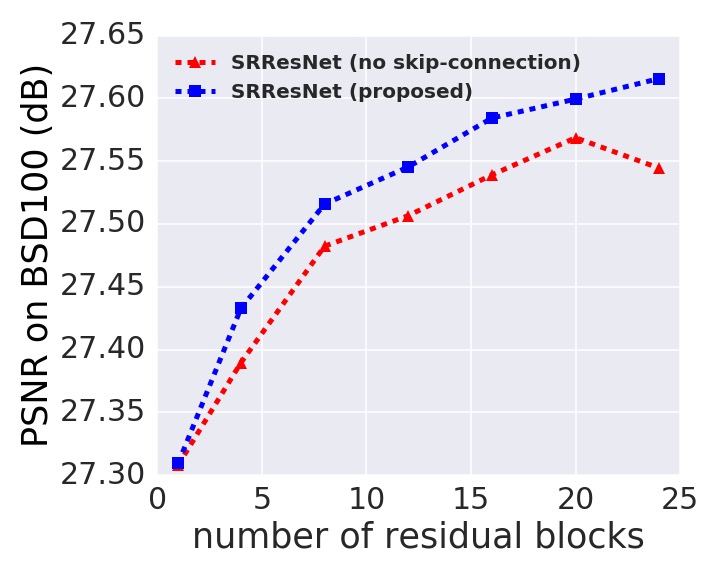} & 
     	     	\includegraphics[width=0.5 \textwidth]{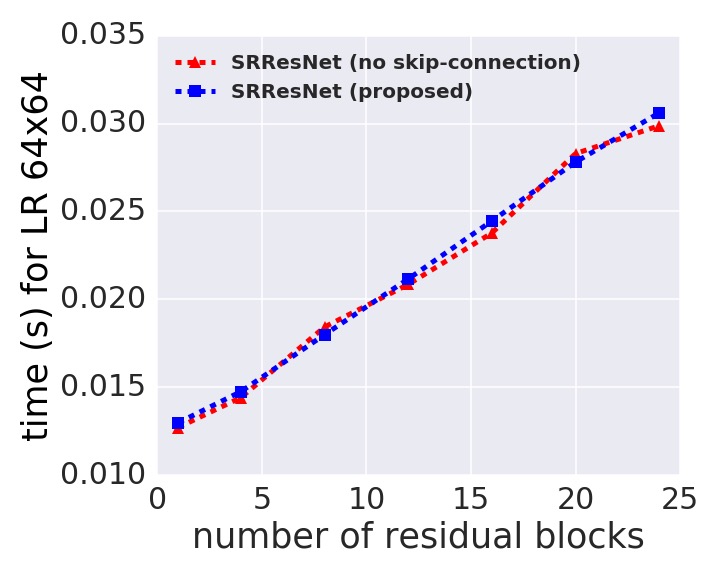}\\
  	\end{tabular}
  		\caption{Dependence of network performance (PSNR, time) on network depth. PSNR (left) calculated on BSD100. Time (right) averaged over 100 reconstructions of a random LR image with resolution 64$\times$64.} 
	\label{fig:app_psnrtime}
\end{figure*}

We investigated the influence of network depth, specifically the number of residual blocks, on performance (PSNR [dB] on BSD100 for $4\times$ SR) and inference time [s] of the network architecture described in Figure 4 of the main paper. Time was assessed on a NVIDIA M40 GPU and averaged over 100 reconstructions of a random low-resolution image with resolution 64$\times$64 with upscaling factor $4\times$. The measurements are plotted in \figurename~\ref{fig:app_psnrtime} for a network with (blue) and without (red) skip-connection.
As expected the time of a single forward pass through the network depends approximately linearly on the number of residual blocks. Whether a skip-connection is used or not has no substantial impact on inference time. However, we observed substantial gains in performance with the additional skip-connection. We chose a network architecture of 16 residual blocks with skip-connection for the evaluation presented in the main paper as we consider this as good trade-off between accuracy and speed including training time. While accuracy gains slowly saturate beyond 16 blocks there is, nevertheless, a clear benefit of using even deeper networks.

\clearpage
\subsection{Evolution of Generator during SRGAN training}
\label{app:evolution}
\begin{figure*}[ht!]
\begin{tabular}{ cx{3.7cm}cx{3.7cm}cx{3.7cm}cx{3.7cm}cx{3.7cm}}
	 ~~~~~~~~~SRResNet & ~~~~~~20k & ~~~~~~~~~~~~~~40k & ~~~~~~~~~~~~~~~~~~60k & ~~~~~~~~~~~~~~~~~~~80k \\
\end{tabular}

\begin{tikzpicture}[zoomboxarray, zoomboxes below, zoomboxarray inner gap=0.1cm, zoomboxarray columns=2, zoomboxarray rows=2]
    \node [image node] { \includegraphics[trim=200 220 40 5, clip, width=1.3in]{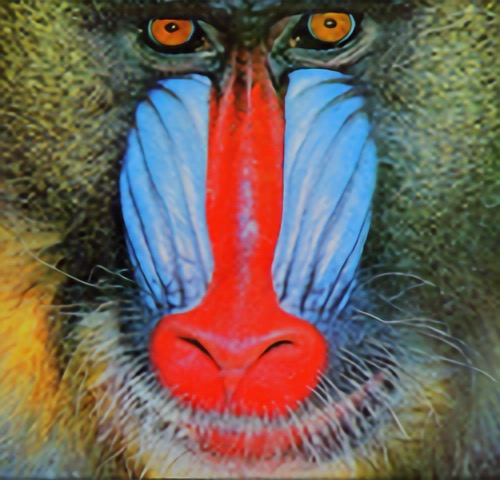} };
    \zoombox[magnification=2,color code=red]{0.5,0.85}
    \zoombox[magnification=2,color code=yellow]{0.7,0.55}
    \zoombox[magnification=2,color code=orange]{0.35,0.3}
    \zoombox[magnification=5,color code=lime]{0.86,0.25}
\end{tikzpicture}
\begin{tikzpicture}[zoomboxarray, zoomboxes below, zoomboxarray inner gap=0.1cm, zoomboxarray columns=2, zoomboxarray rows=2]
    \node [image node] { \includegraphics[trim=200 220 40 5, clip, width=1.3in]{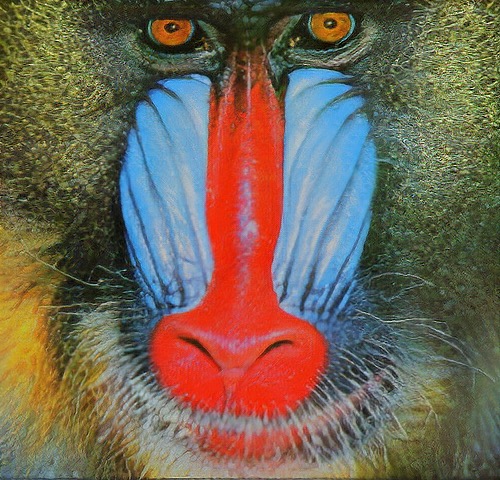} };
    \zoombox[magnification=2,color code=red]{0.5,0.85}
    \zoombox[magnification=2,color code=yellow]{0.7,0.55}
    \zoombox[magnification=2,color code=orange]{0.35,0.3}
    \zoombox[magnification=5,color code=lime]{0.86,0.25}
\end{tikzpicture}
\begin{tikzpicture}[zoomboxarray, zoomboxes below, zoomboxarray inner gap=0.1cm, zoomboxarray columns=2, zoomboxarray rows=2]
    \node [image node] { \includegraphics[trim=200 220 40 5, clip, width=1.3in]{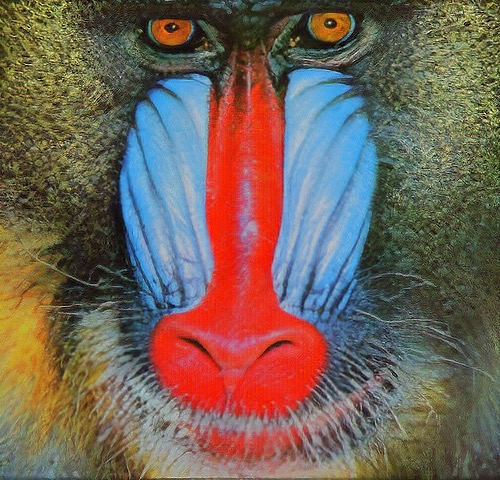} };
    \zoombox[magnification=2,color code=red]{0.5,0.85}
    \zoombox[magnification=2,color code=yellow]{0.7,0.55}
    \zoombox[magnification=2,color code=orange]{0.35,0.3}
    \zoombox[magnification=5,color code=lime]{0.86,0.25}
\end{tikzpicture}
\begin{tikzpicture}[zoomboxarray, zoomboxes below, zoomboxarray inner gap=0.1cm, zoomboxarray columns=2, zoomboxarray rows=2]
    \node [image node] { \includegraphics[trim=200 220 40 5, clip, width=1.3in]{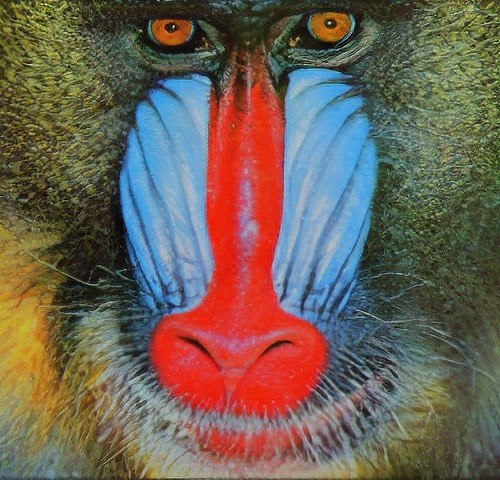} };
    \zoombox[magnification=2,color code=red]{0.5,0.85}
    \zoombox[magnification=2,color code=yellow]{0.7,0.55}
    \zoombox[magnification=2,color code=orange]{0.35,0.3}
    \zoombox[magnification=5,color code=lime]{0.86,0.25}
\end{tikzpicture}
\begin{tikzpicture}[zoomboxarray, zoomboxes below, zoomboxarray inner gap=0.1cm, zoomboxarray columns=2, zoomboxarray rows=2]
    \node [image node] { \includegraphics[trim=200 220 40 5, clip, width=1.3in]{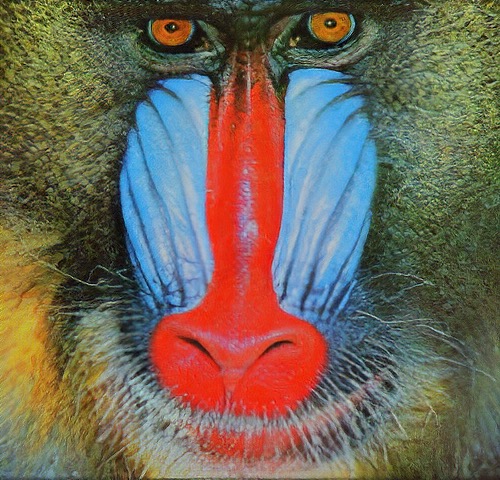} };
    \zoombox[magnification=2,color code=red]{0.5,0.85}
    \zoombox[magnification=2,color code=yellow]{0.7,0.55}
    \zoombox[magnification=2,color code=orange]{0.35,0.3}
    \zoombox[magnification=5,color code=lime]{0.86,0.25}
\end{tikzpicture}\\
\begin{tabular}{ cx{3.7cm}cx{3.7cm}cx{3.7cm}cx{3.7cm}cx{3.7cm}}
	 ~~~~~~~~~~~~100k & ~~~~~~~~~~~~~~~~~140k & ~~~~~~~~~~~~~~~~~180k & ~~~~~~~~~~~~~~~~~~~SRGAN & ~~~~~~~~original HR image \\
\end{tabular}
\begin{tikzpicture}[zoomboxarray, zoomboxes below, zoomboxarray inner gap=0.1cm, zoomboxarray columns=2, zoomboxarray rows=2]
    \node [image node] { \includegraphics[trim=200 220 40 5, clip, width=1.3in]{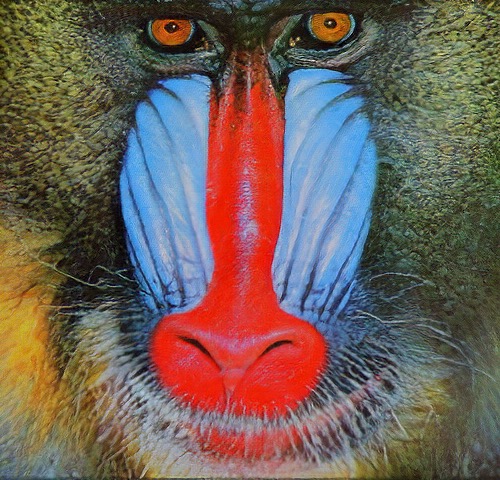} };
    \zoombox[magnification=2,color code=red]{0.5,0.85}
    \zoombox[magnification=2,color code=yellow]{0.7,0.55}
    \zoombox[magnification=2,color code=orange]{0.35,0.3}
    \zoombox[magnification=5,color code=lime]{0.86,0.25}
\end{tikzpicture}
\begin{tikzpicture}[zoomboxarray, zoomboxes below, zoomboxarray inner gap=0.1cm, zoomboxarray columns=2, zoomboxarray rows=2]
    \node [image node] { \includegraphics[trim=200 220 40 5, clip, width=1.3in]{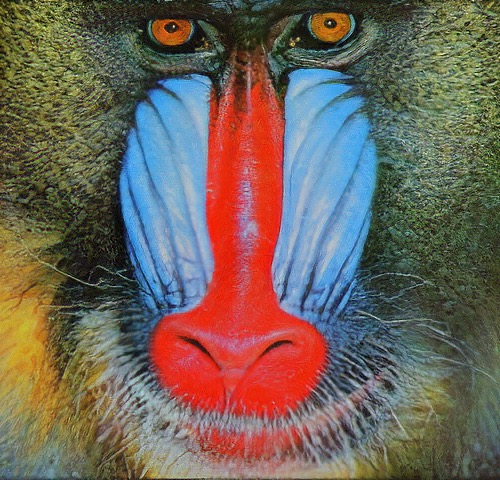} };
    \zoombox[magnification=2,color code=red]{0.5,0.85}
    \zoombox[magnification=2,color code=yellow]{0.7,0.55}
    \zoombox[magnification=2,color code=orange]{0.35,0.3}
    \zoombox[magnification=5,color code=lime]{0.86,0.25}
\end{tikzpicture}
\begin{tikzpicture}[zoomboxarray, zoomboxes below, zoomboxarray inner gap=0.1cm, zoomboxarray columns=2, zoomboxarray rows=2]
    \node [image node] { \includegraphics[trim=200 220 40 5, clip, width=1.3in]{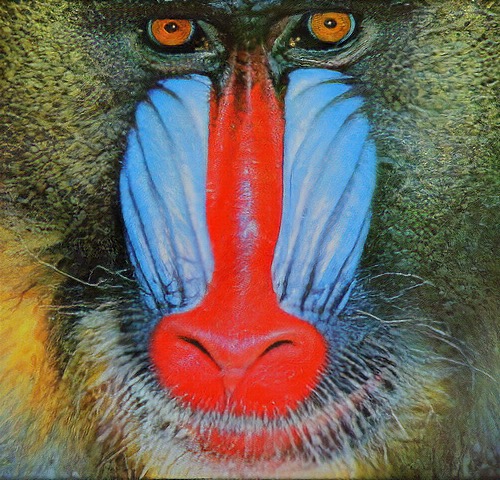} };
    \zoombox[magnification=2,color code=red]{0.5,0.85}
    \zoombox[magnification=2,color code=yellow]{0.7,0.55}
    \zoombox[magnification=2,color code=orange]{0.35,0.3}
    \zoombox[magnification=5,color code=lime]{0.86,0.25}
\end{tikzpicture}
\begin{tikzpicture}[zoomboxarray, zoomboxes below, zoomboxarray inner gap=0.1cm, zoomboxarray columns=2, zoomboxarray rows=2]
    \node [image node] { \includegraphics[trim=200 220 40 5, clip, width=1.3in]{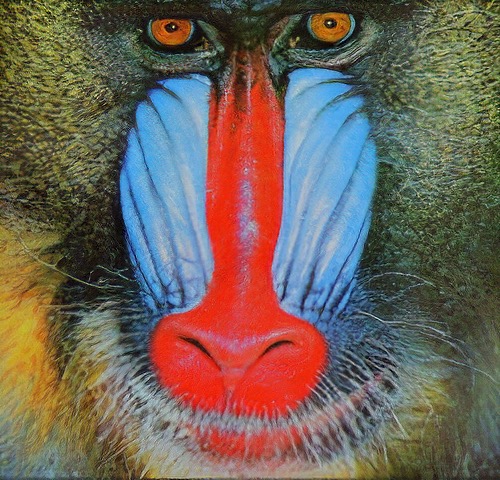} };
    \zoombox[magnification=2,color code=red]{0.5,0.85}
    \zoombox[magnification=2,color code=yellow]{0.7,0.55}
    \zoombox[magnification=2,color code=orange]{0.35,0.3}
    \zoombox[magnification=5,color code=lime]{0.86,0.25}
\end{tikzpicture}
\begin{tikzpicture}[zoomboxarray, zoomboxes below, zoomboxarray inner gap=0.1cm, zoomboxarray columns=2, zoomboxarray rows=2]
    \node [image node] { \includegraphics[trim=200 220 40 5, clip, width=1.3in]{images/used/jpg/baboon_HR} };
    \zoombox[magnification=2,color code=red]{0.5,0.85}
    \zoombox[magnification=2,color code=yellow]{0.7,0.55}
    \zoombox[magnification=2,color code=orange]{0.35,0.3}
    \zoombox[magnification=5,color code=lime]{0.86,0.25}
\end{tikzpicture}

\caption{Evolution of SRGAN generator network during training progress. Note: Generator initialized with SRResNet weights; learning rate set to  $10^{-4}$ for first 100k iterations, then reduced to $10^{-5}$ for another 100k iterations. [$4\times$ upscaling]}
\label{fig:app_perceptual}
\end{figure*}

We further investigated how reconstructions of the \textbf{SRGAN} generator network evolve (visually) with increasing number of training iterations. Visual results obtained after different number of training iterations are illustrated in \figurename~\ref{fig:app_perceptual}. It is interesting that after only 20 thousand training iterations the generator substantially diverged from the SRResNet initialization and produces reconstruction with a lot of high frequency content, including noise. With increasing number of training iterations reconstructions of the \textit{baboon} from \textbf{Set14} appear closer to the reference image. However, there is visually little change during the last 50-100 thousand update iterations.

\clearpage

\subsection{Mean opinion score (MOS) testing}
\label{app:MOS}

\begin{figure*}[h!]
  	\begin{tabular}{ccc}
  		 Set5 & Set14 & BSD100 \\
     	\includegraphics[width=0.3\textwidth]{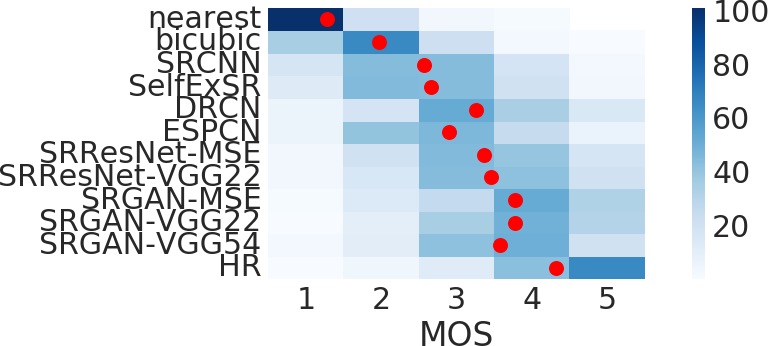}&
     	\includegraphics[width=0.3\textwidth]{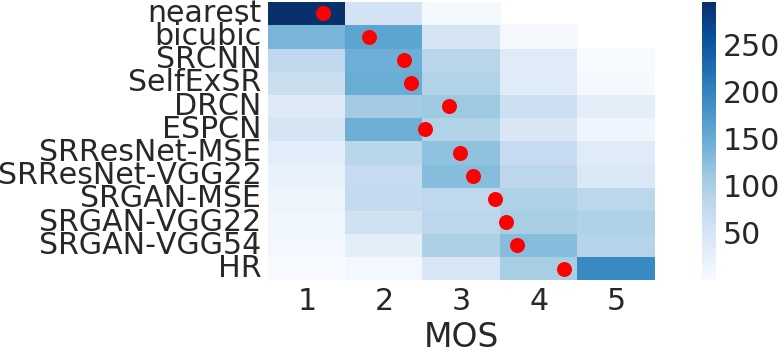} &
     	\includegraphics[width=0.3\textwidth]{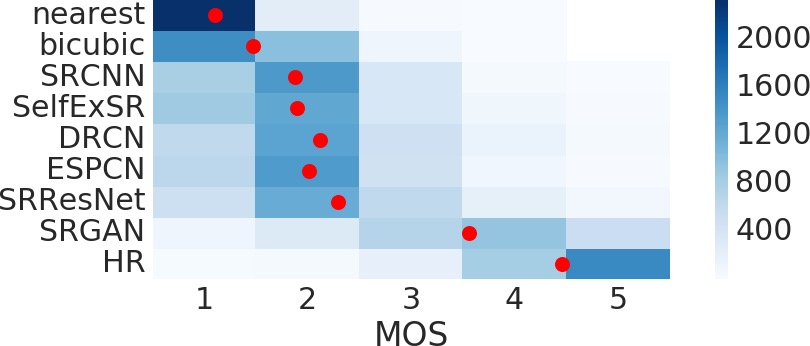} \\
     \end{tabular}
     \caption{Color-coded distribution of MOS scores on Set5, Set14, BSD100. Mean shown as red marker, where the bins are centered around value i. [$4\times$ upscaling]}
     \label{fig:app_mosdist}
\end{figure*}

\begin{figure*}[h!]
  	\begin{tabular}{ccc}
  		 Set5 & Set14 & BSD100 \\
     	\includegraphics[width=0.3\textwidth]{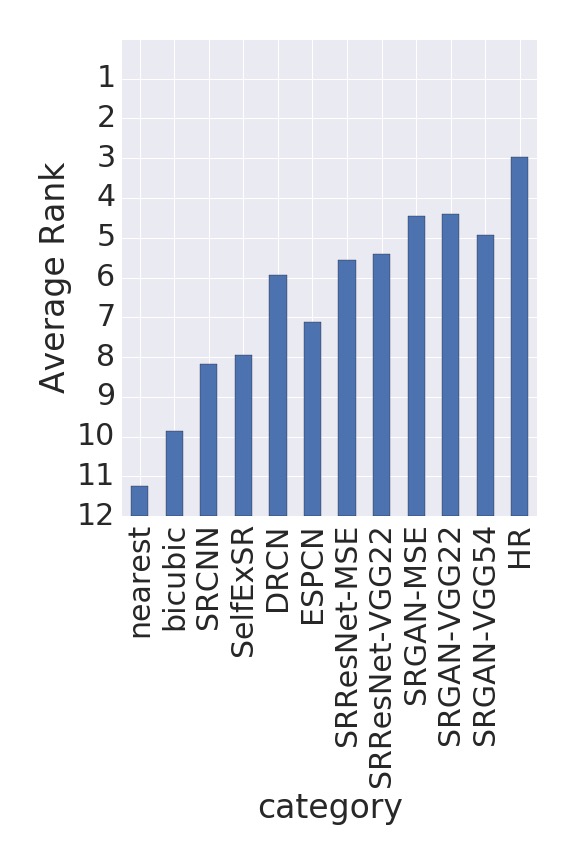}&
     	\includegraphics[width=0.3\textwidth]{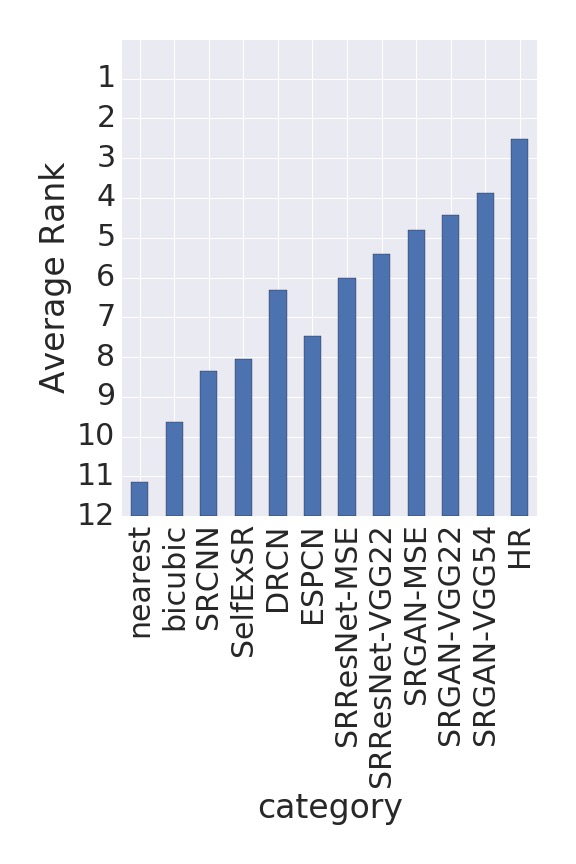} &
     	\includegraphics[width=0.3\textwidth]{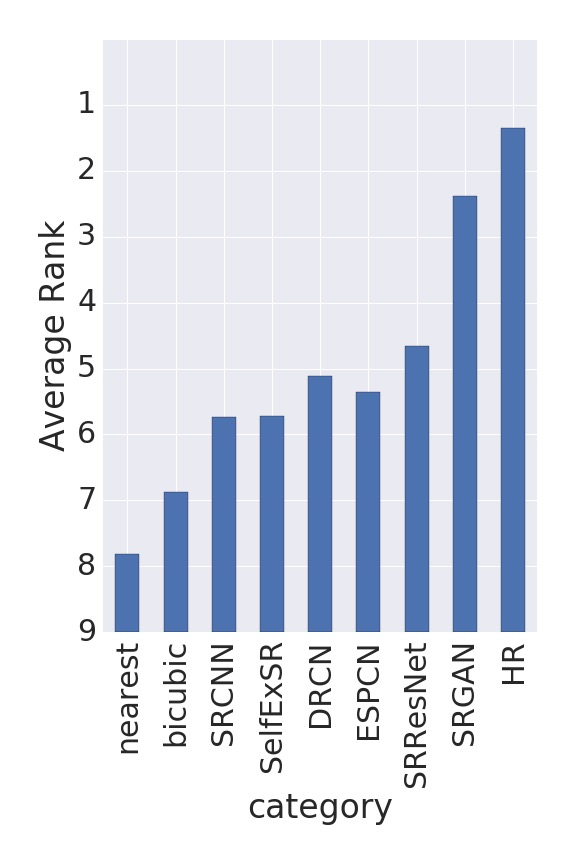} \\
     \end{tabular}
     \caption{Average rank on Set5, Set14, BSD100 by averaging the ranks over all available individual ratings. [$4\times$ upscaling]}
     \label{fig:app_mosrank}
\end{figure*}

In all conducted \ac{MOS} tests we have asked 26 human raters to assign a score from 1 (Bad) to 5 (Excellent) to reconstructions of the $4\times$ downsampled versions of images from \textbf{Set5}, \textbf{Set14} and \textbf{BSD100}. On \textbf{BSD100} nine versions of each image were rated by each rater. On \textbf{Set5} and \textbf{Set14} the raters also rated three additional versions of the proposed methods to investigate different content losses. In total 26*100*9 + 26*14*12 + 26*5*12 = 29328 ratings were obtained, where each rater rated 1128 images. Images were presented in a completely randomized fashion without any indication of the employed super-resolution approach. The raters were calibrated on images not included in the testing set such that the nearest neighbor interpolated reconstruction should receive score 1 (Bad) and the original high-resolution image score 5 (Excellent). 
The distribution of \ac{MOS} ratings on each individual data set is summarized in \figurename~\ref{fig:app_mosdist}. The average ordinal rank over all corresponding ratings of an image and rater are shown in \figurename~\ref{fig:app_mosrank}. Note that a score of 1 corresponds to the best rank and ranks are averaged for samples that would have the same ordinal ranking.
While results on \textbf{Set5} are somewhat inconclusive due to very small sample size and images with comparably little detail, ratings on \textbf{Set14} and especially on the large \textbf{BSD100} data set confirm that \textbf{SRGAN} is significantly better than any compared state-of-the-art method. In fact, \ac{MOS} ratings obtained with \textbf{SRGAN} are closer to those of the original high-resolution images than to those obtained with any reference method. 
\clearpage
\subsection{Set5 - Visual Results}
\label{app:Set5}
\begin{figure*}[h!]
  	\begin{tabular}{cccc}
  		 bicubic & SRResNet & SRGAN & original \\
     	\includegraphics[width=1.4in]{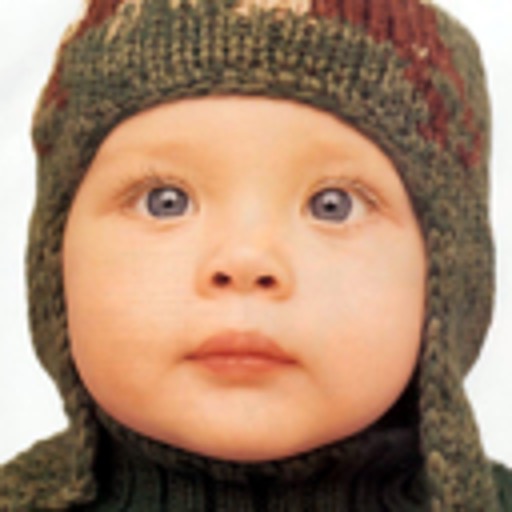}&
     	\includegraphics[width=1.4in]{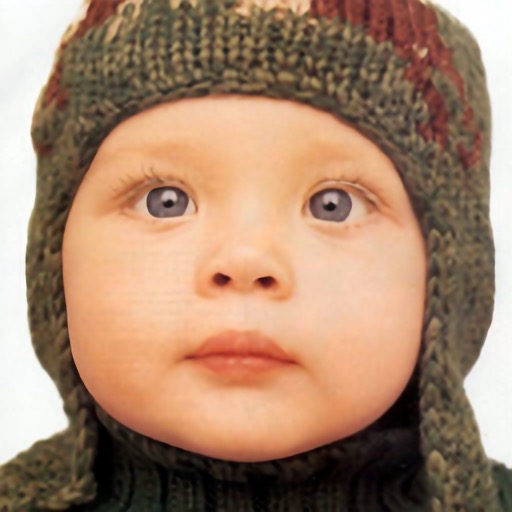} &
     	\includegraphics[width=1.4in]{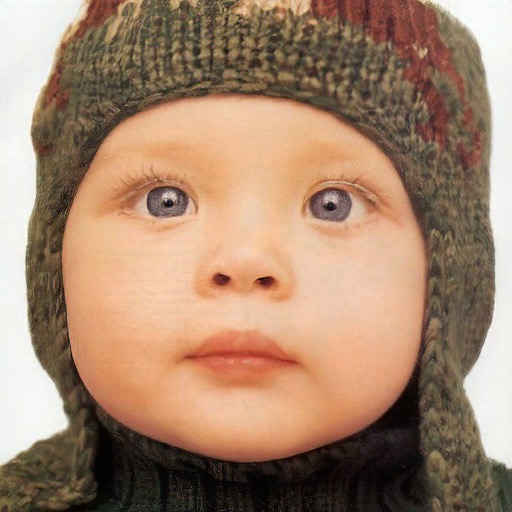} &
     	\includegraphics[width=1.4in]{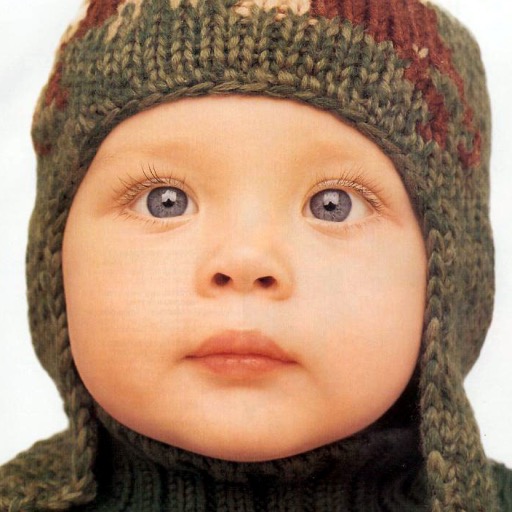} \\
     	\includegraphics[width=1.4in]{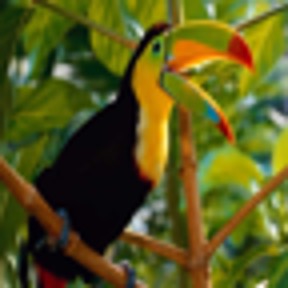}&
     	\includegraphics[width=1.4in]{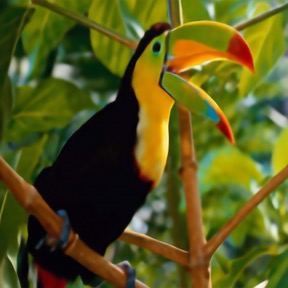} &
     	\includegraphics[width=1.4in]{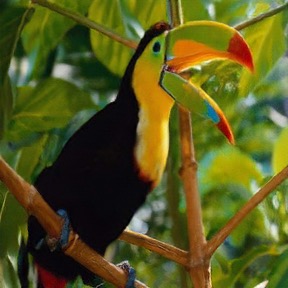} &
     	\includegraphics[width=1.4in]{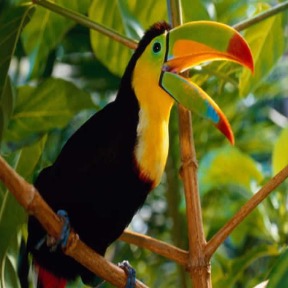} \\
     	\includegraphics[width=1.4in]{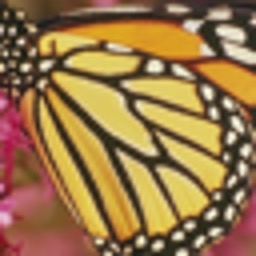}&
     	\includegraphics[width=1.4in]{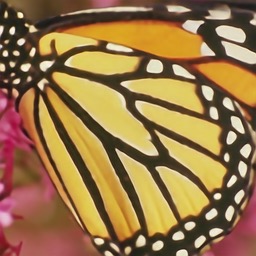} &
     	\includegraphics[width=1.4in]{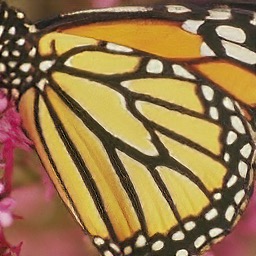} &
     	\includegraphics[width=1.4in]{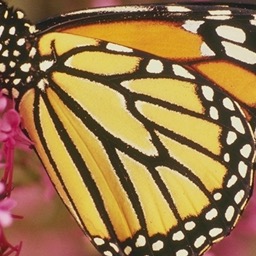} \\
       	\includegraphics[width=1.4in]{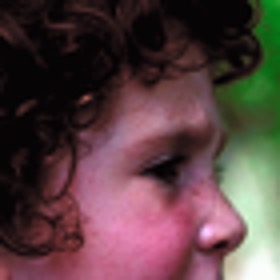}&
     	\includegraphics[width=1.4in]{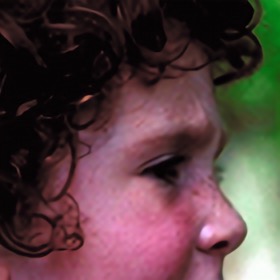} &
     	\includegraphics[width=1.4in]{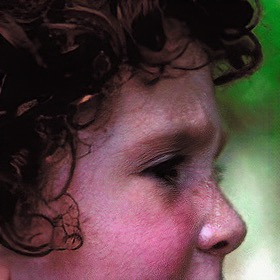} &
     	\includegraphics[width=1.4in]{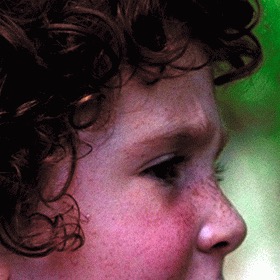} \\
     	\includegraphics[width=1.4in]{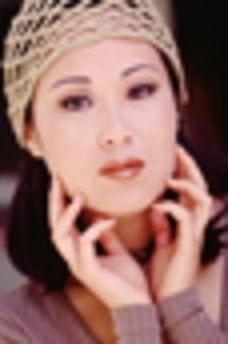}&
     	\includegraphics[width=1.4in]{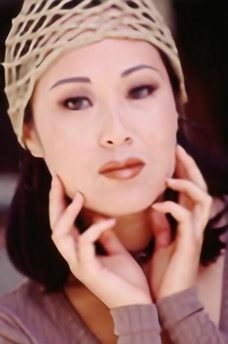} &
     	\includegraphics[width=1.4in]{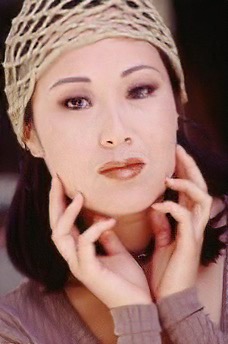} &
     	\includegraphics[width=1.4in]{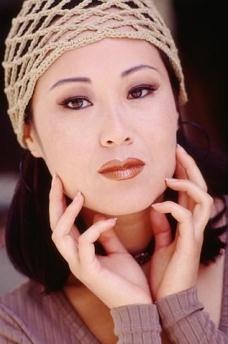} \\
  	\end{tabular}
  	\label{fig:app_Set5}
  	\caption{Results for \textbf{Set5} using bicubic interpolation, SRResNet and SRGAN. [$4\times$ upscaling]} 
\end{figure*}

\clearpage

\subsection{Set14 - Visual Results}
\label{app:Set14}
\begin{figure*}[h!] 
  	\begin{tabular}{cccc}
  		bicubic & SRResNet & SRGAN & original \\
     	\includegraphics[width=1.5in]{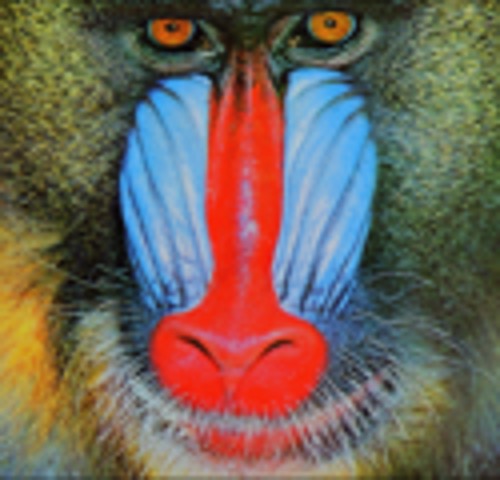}&
     	\includegraphics[width=1.5in]{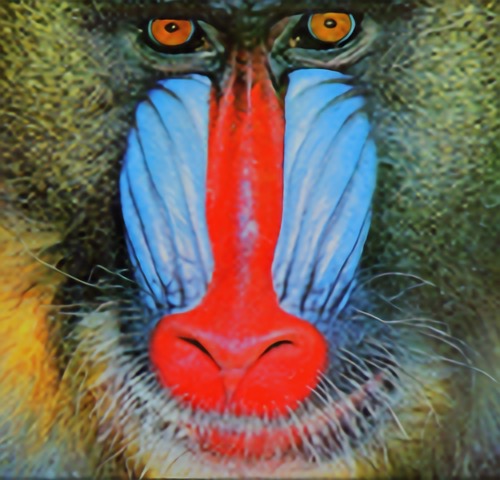} &
     	\includegraphics[width=1.5in]{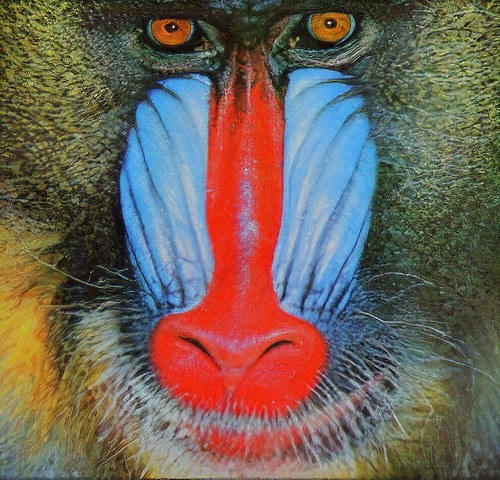} &
     	\includegraphics[width=1.5in]{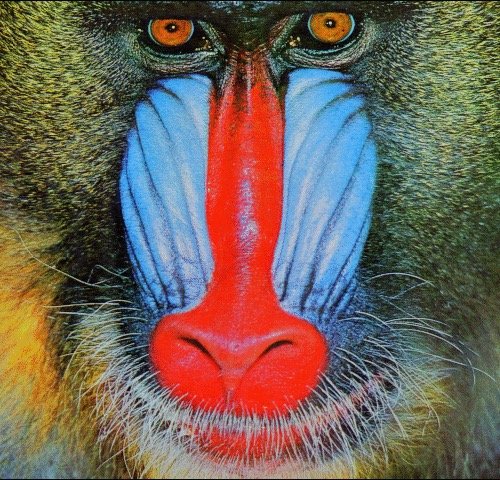} \\
     	\includegraphics[width=1.5in]{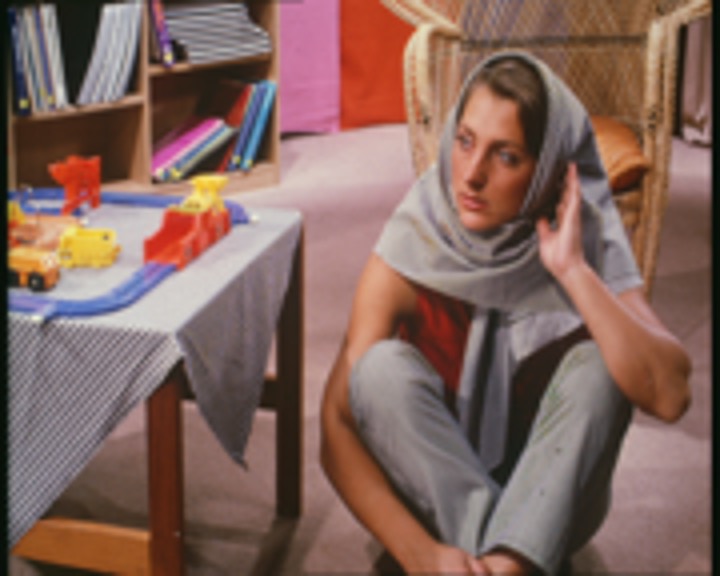}&
     	\includegraphics[width=1.5in]{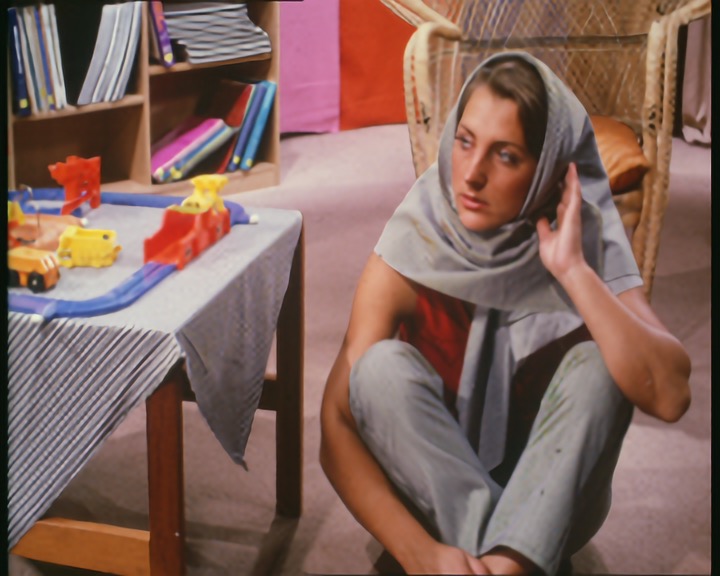} &
     	\includegraphics[width=1.5in]{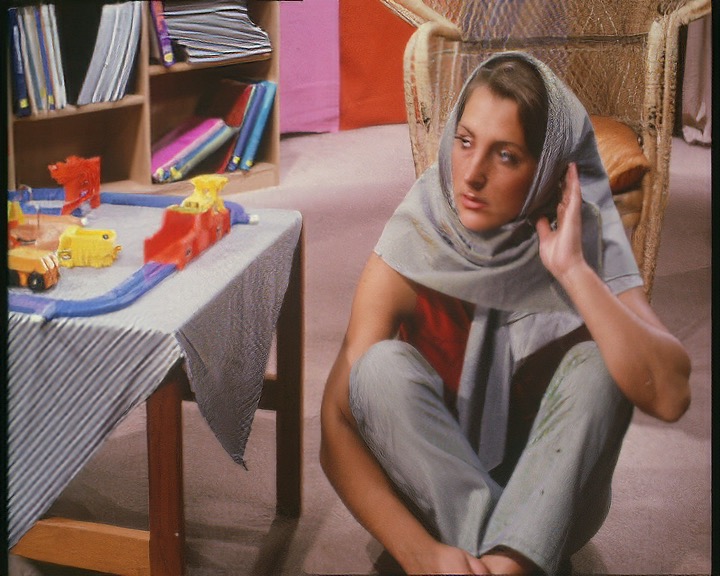} &
     	\includegraphics[width=1.5in]{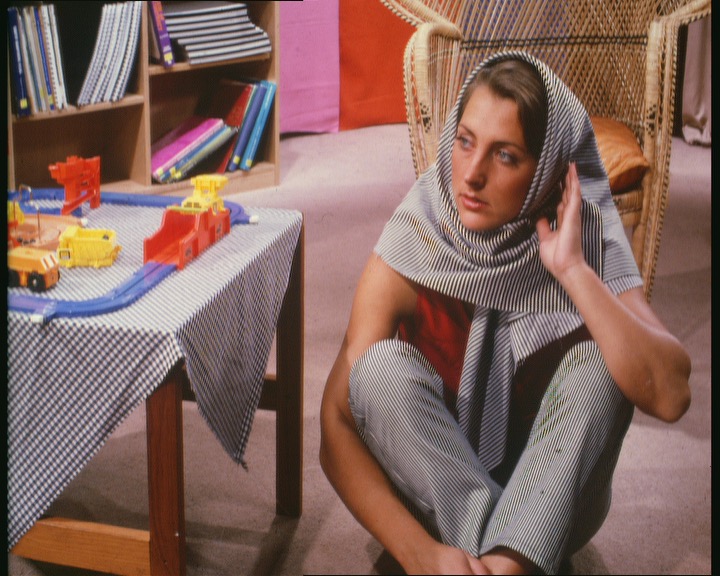} \\
     	\includegraphics[width=1.5in]{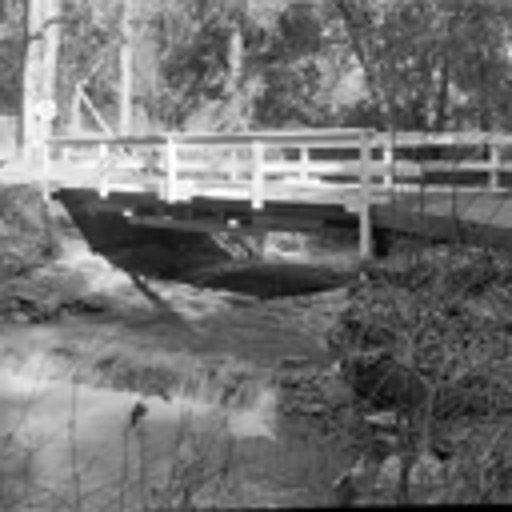}&
     	\includegraphics[width=1.5in]{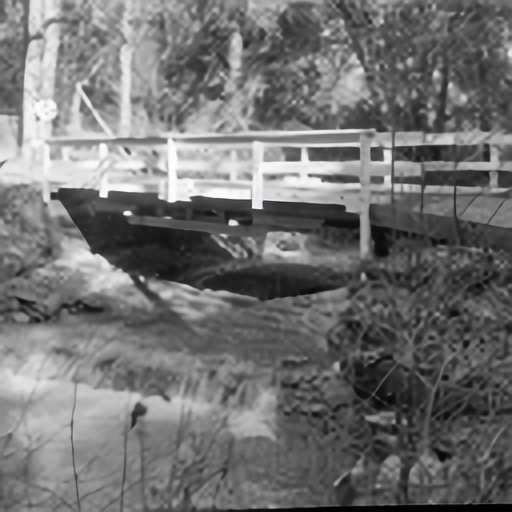} &
     	\includegraphics[width=1.5in]{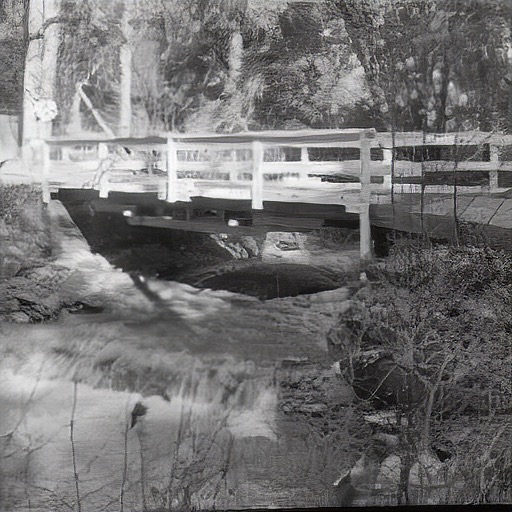} &
     	\includegraphics[width=1.5in]{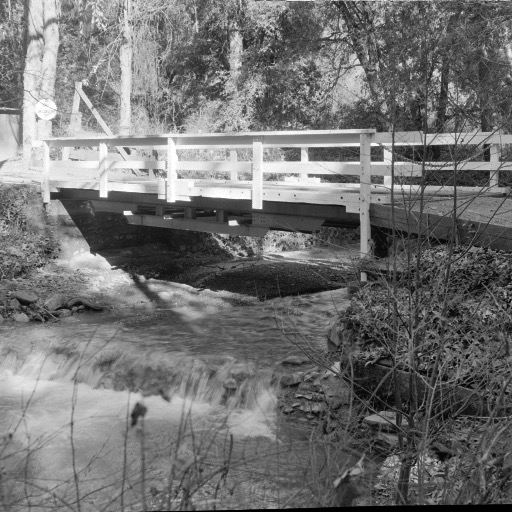} \\
     	\includegraphics[width=1.5in]{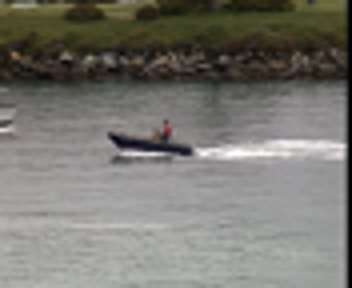}&
     	\includegraphics[width=1.5in]{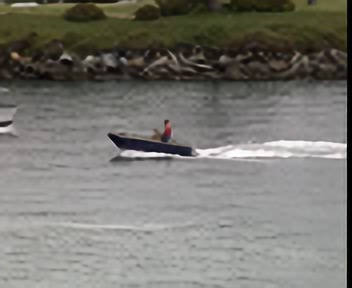} &
     	\includegraphics[width=1.5in]{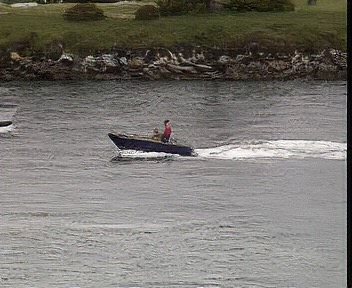} &
     	\includegraphics[width=1.5in]{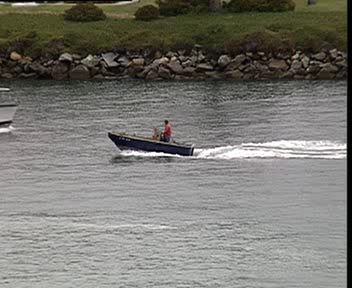} \\
     	\includegraphics[width=1.5in]{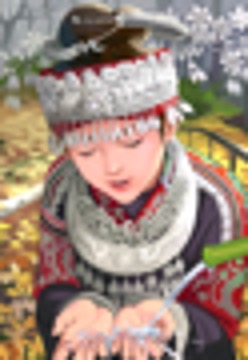}&
     	\includegraphics[width=1.5in]{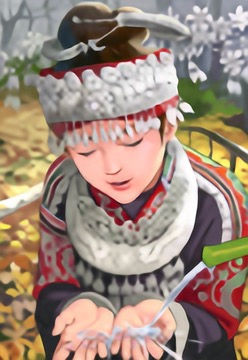} &
     	\includegraphics[width=1.5in]{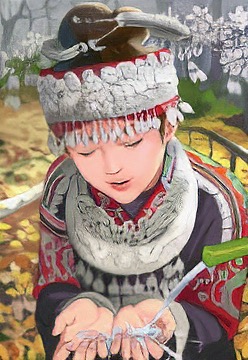} &
     	\includegraphics[width=1.5in]{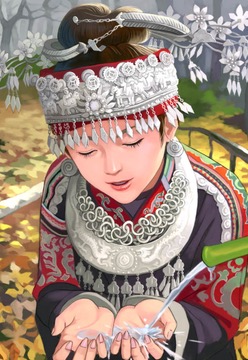} \\     	
  	\end{tabular}
  	\label{fig:app_Set14a}
  	\caption{Results for \textbf{Set14} using bicubic interpolation, SRResNet and SRGAN. [$4\times$ upscaling]} 
\end{figure*}

\begin{figure*}[h!] 
  	\begin{tabular}{cccc}
  		 bicubic & SRResNet & SRGAN & original \\
     	\includegraphics[width=1.5in]{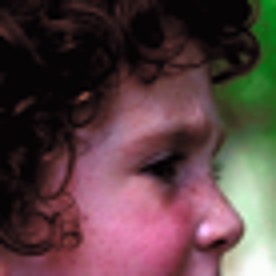}&
     	\includegraphics[width=1.5in]{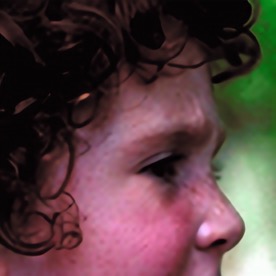} &
     	\includegraphics[width=1.5in]{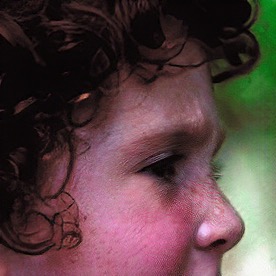} &
     	\includegraphics[width=1.5in]{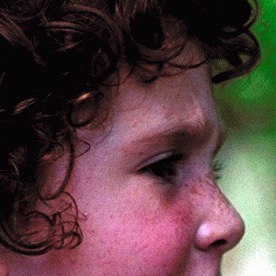} \\
     	\includegraphics[width=1.5in]{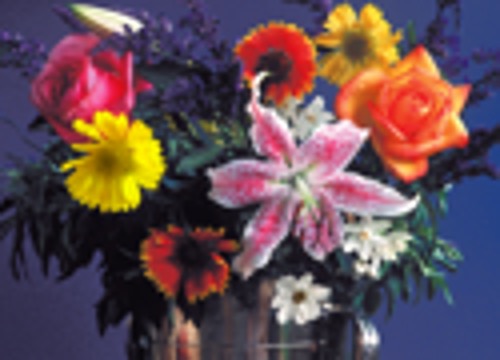}&
     	\includegraphics[width=1.5in]{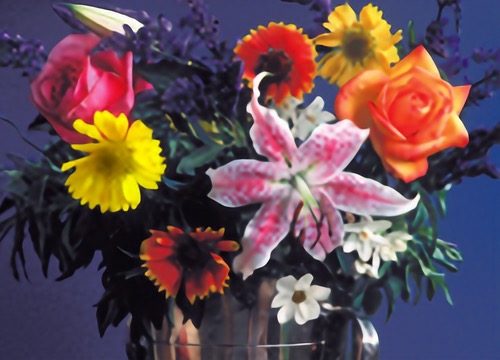} &
     	\includegraphics[width=1.5in]{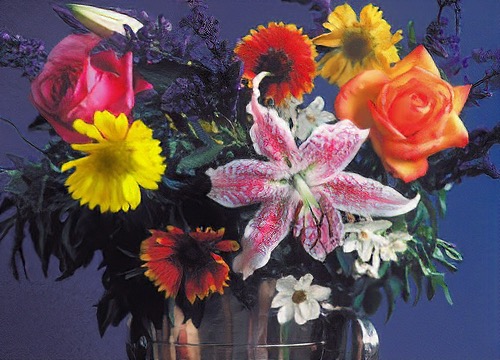} &	
     	\includegraphics[width=1.5in]{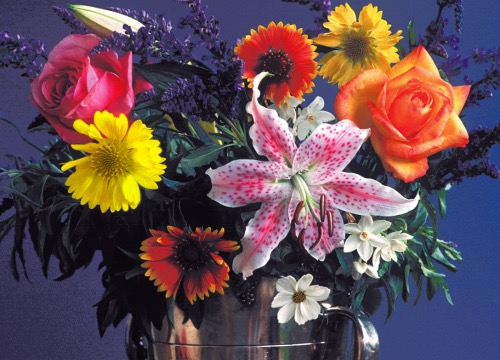} \\
     	\includegraphics[width=1.5in]{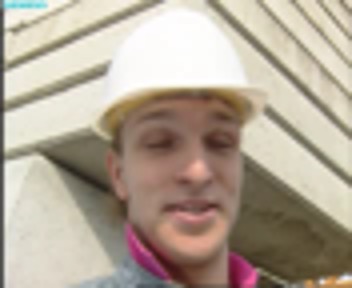}&
     	\includegraphics[width=1.5in]{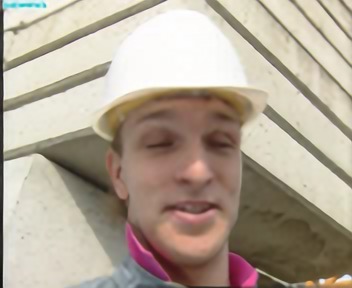} &
     	\includegraphics[width=1.5in]{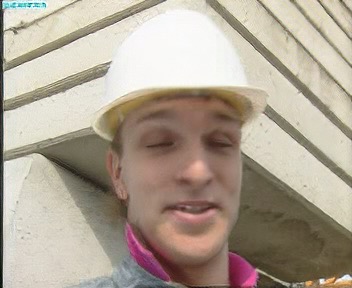} &
     	\includegraphics[width=1.5in]{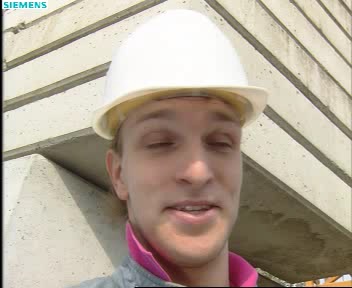} \\
     	\includegraphics[width=1.5in]{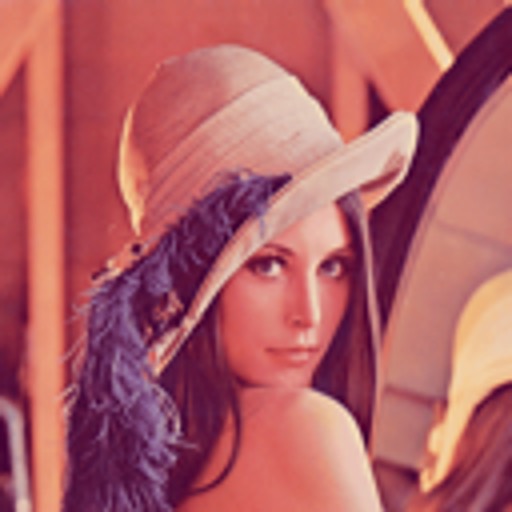}&
     	\includegraphics[width=1.5in]{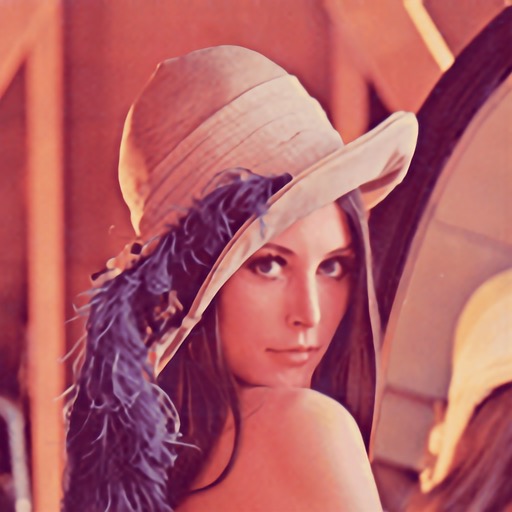} &
     	\includegraphics[width=1.5in]{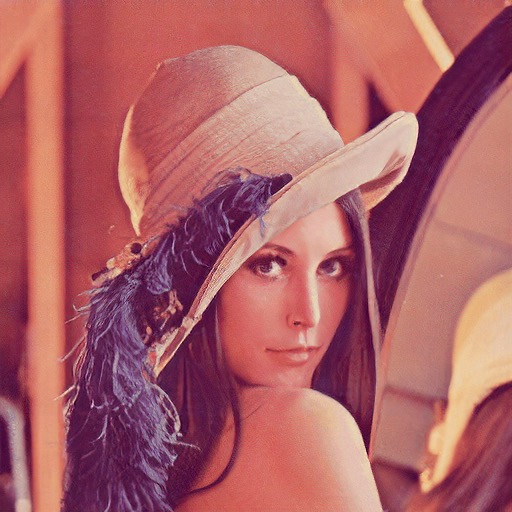} & 
     	\includegraphics[width=1.5in]{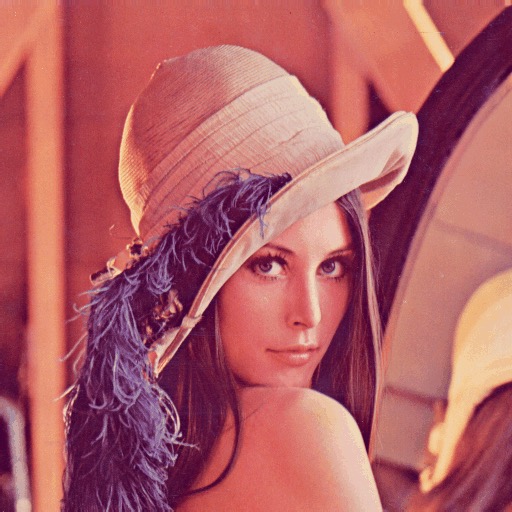} \\ 	
     	\includegraphics[width=1.5in]{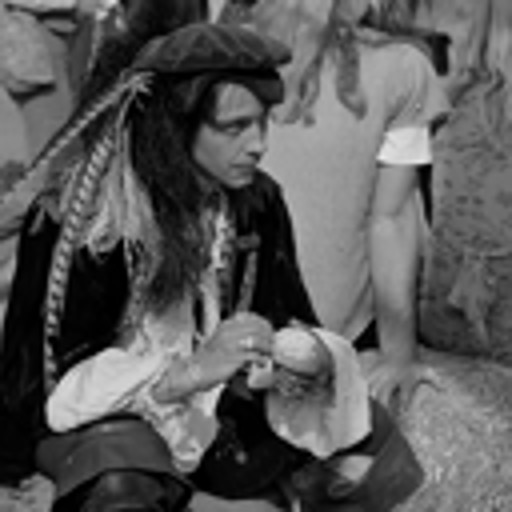}&
     	\includegraphics[width=1.5in]{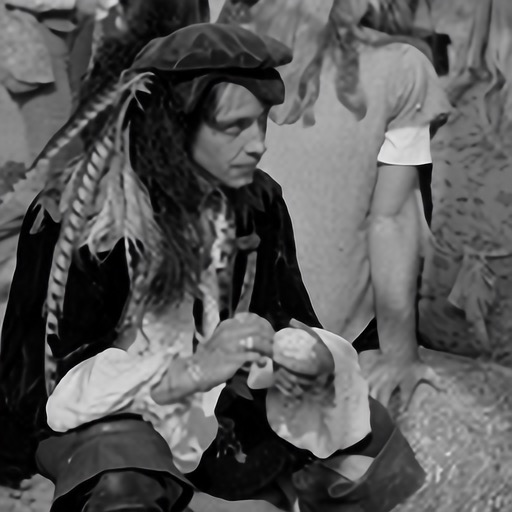} &
     	\includegraphics[width=1.5in]{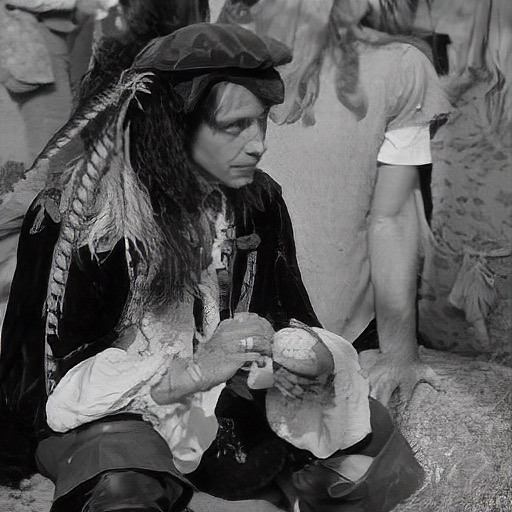} &
     	\includegraphics[width=1.5in]{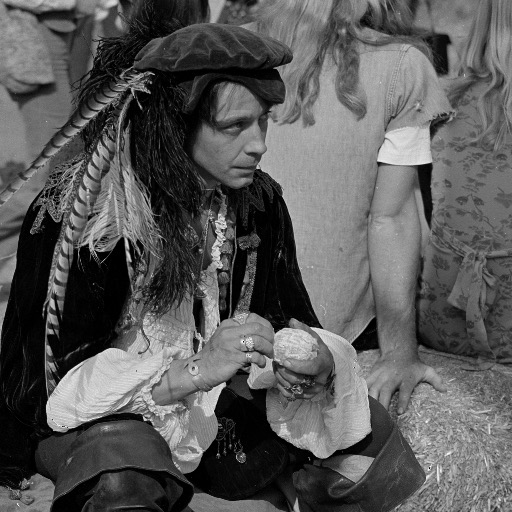} \\     	
  	\end{tabular}
  	\label{fig:app_Set14b}
  	\caption{Results for \textbf{Set14} using bicubic interpolation , SRResNet and SRGAN. [$4\times$ upscaling]} 
\end{figure*}

\begin{figure*}[h!] 
  	\begin{tabular}{cccc}
  		 bicubic & SRResNet & SRGAN & original \\
     	\includegraphics[width=1.5in]{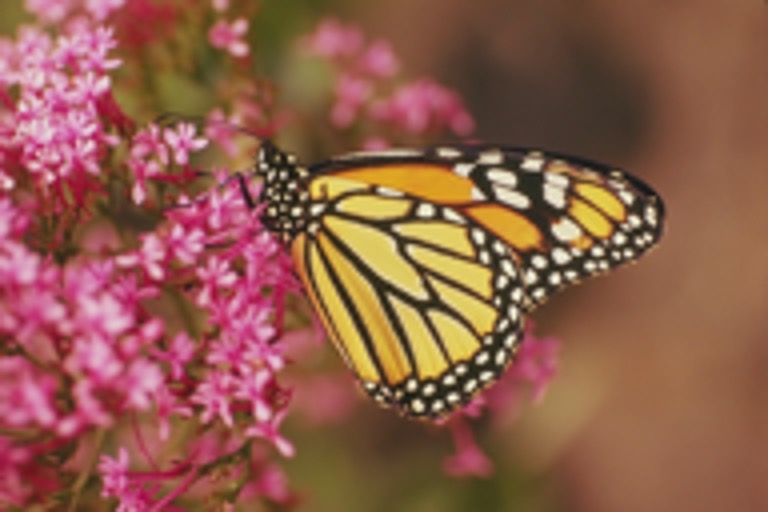}&
     	\includegraphics[width=1.5in]{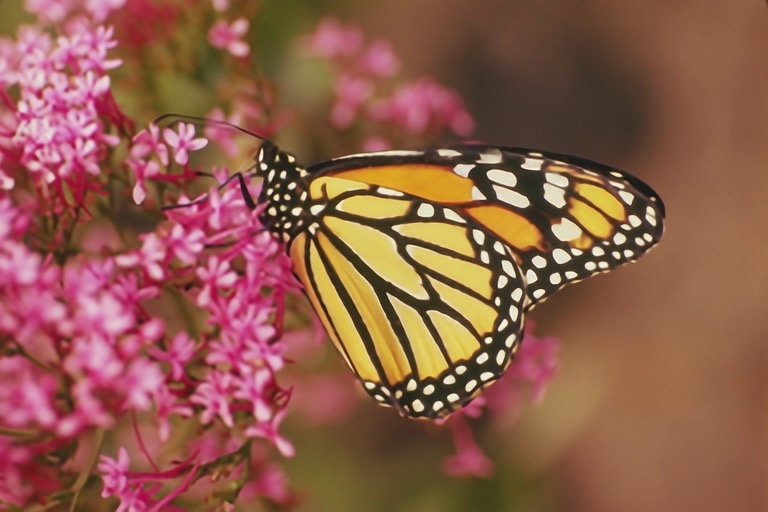} &
     	\includegraphics[width=1.5in]{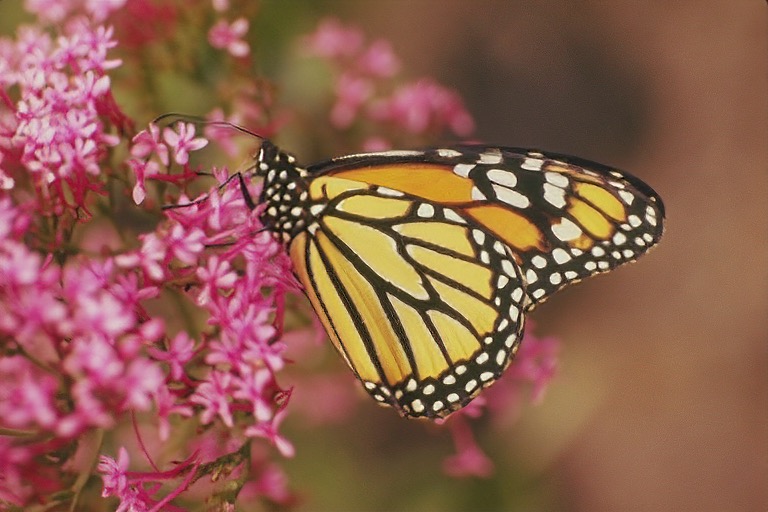} &
     	\includegraphics[width=1.5in]{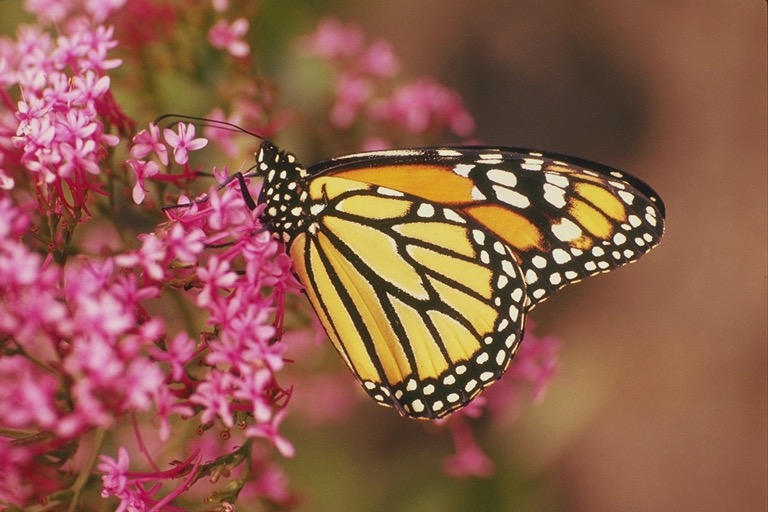} \\     	
     	\includegraphics[width=1.5in]{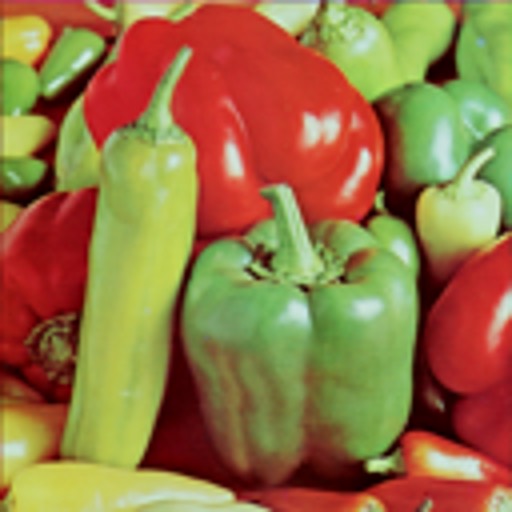}&
     	\includegraphics[width=1.5in]{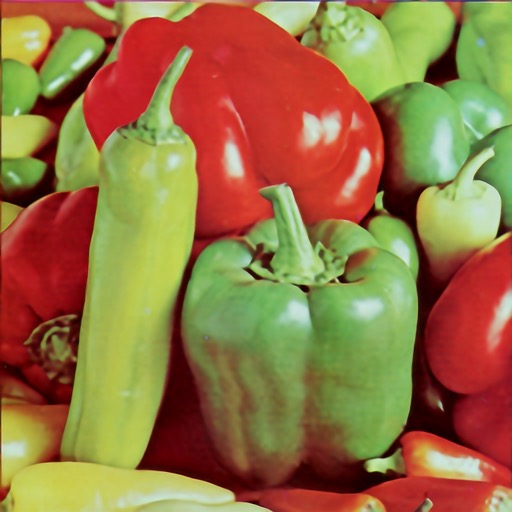} &
     	\includegraphics[width=1.5in]{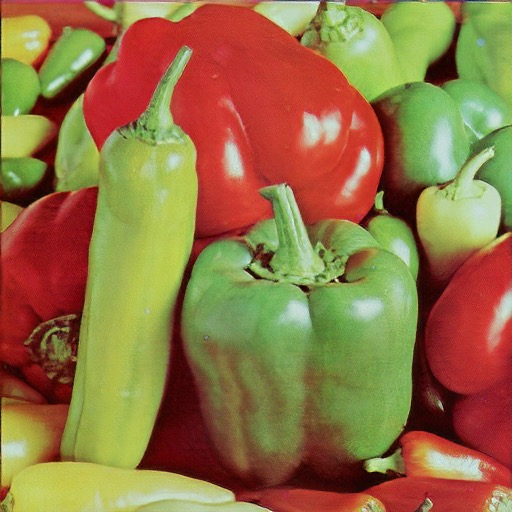} &
     	\includegraphics[width=1.5in]{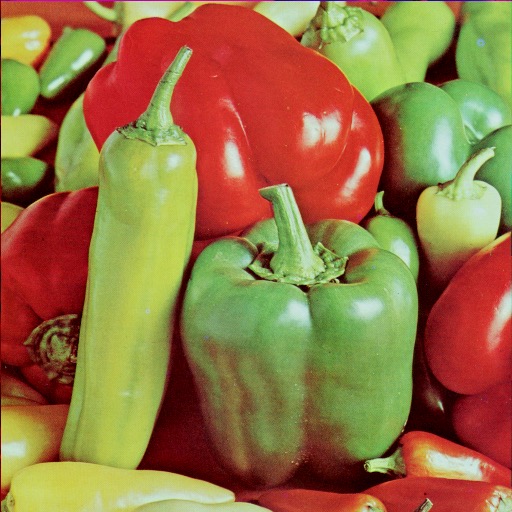} \\     	
     	\includegraphics[width=1.5in]{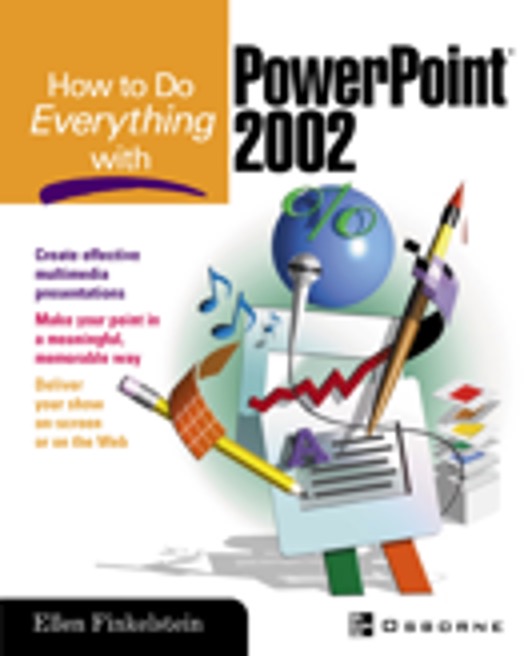}&
     	\includegraphics[width=1.5in]{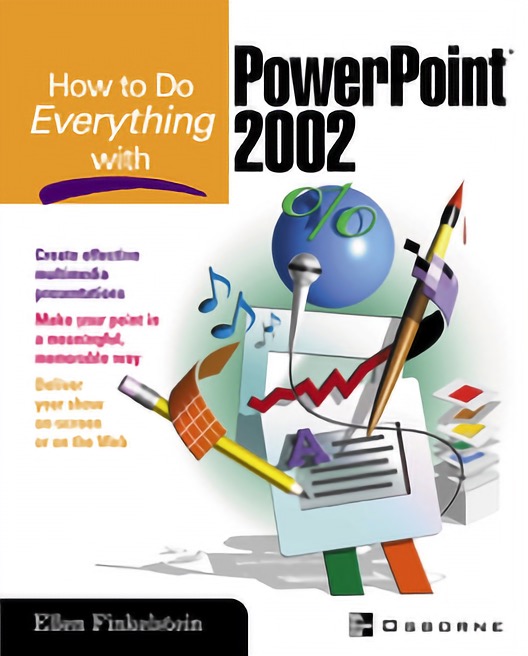} &
     	\includegraphics[width=1.5in]{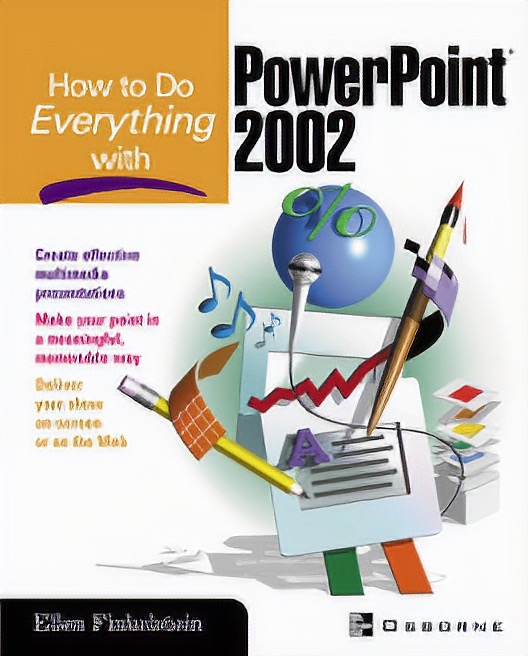} &
     	\includegraphics[width=1.5in]{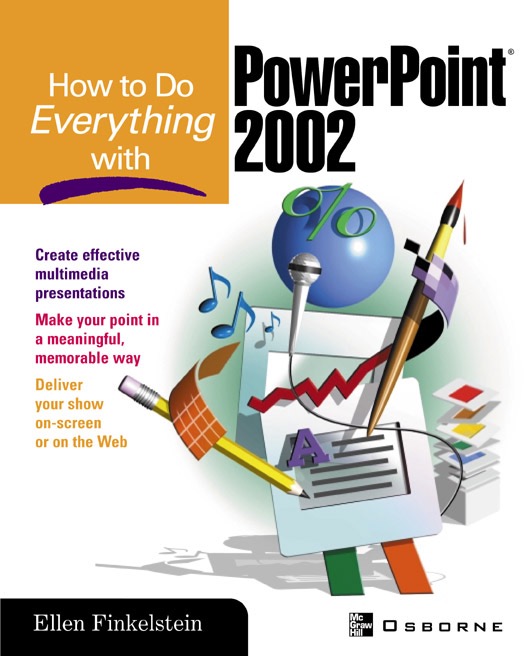} \\     	
     	\includegraphics[width=1.5in]{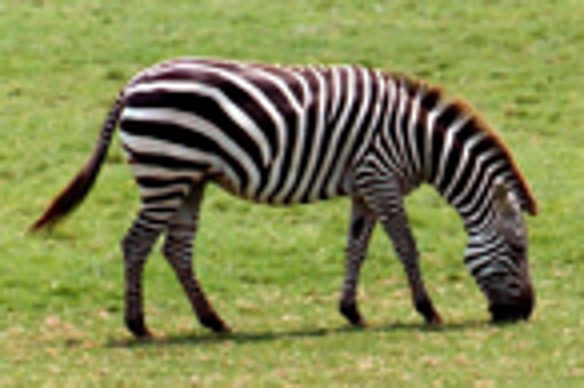}&
     	\includegraphics[width=1.5in]{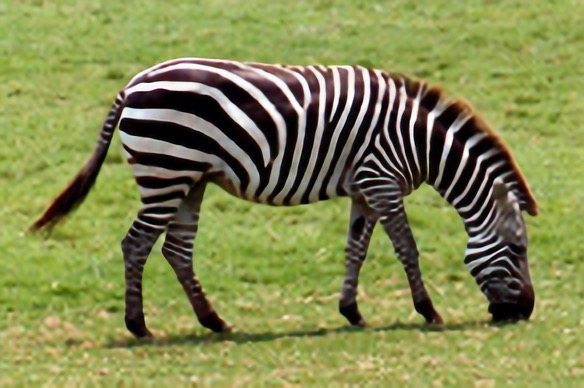} &
     	\includegraphics[width=1.5in]{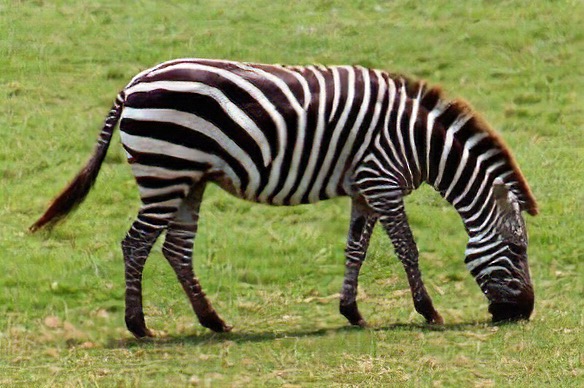} &   
     	\includegraphics[width=1.5in]{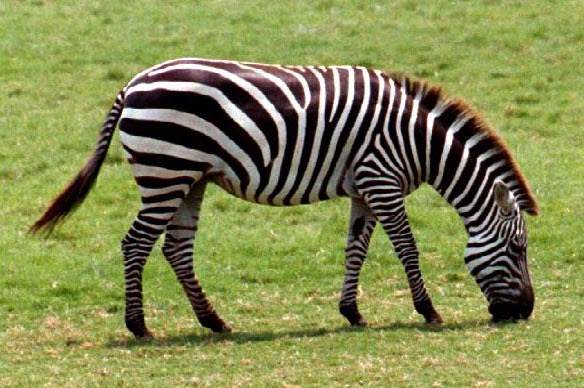} \\     	  	     
  	\end{tabular}
  	\label{fig:app_Set14c}
  	\caption{Results for \textbf{Set14} using bicubic interpolation, SRResNet and SRGAN. [$4\times$ upscaling]} 
\end{figure*}
\clearpage
\subsection{BSD100 (five random samples) - Visual Results}
\label{app:BSD100}
\begin{figure*}[h!] 
  	\begin{tabular}{cccc}
  		 bicubic & SRResNet & SRGAN & original \\
     	\includegraphics[width=1.5in]{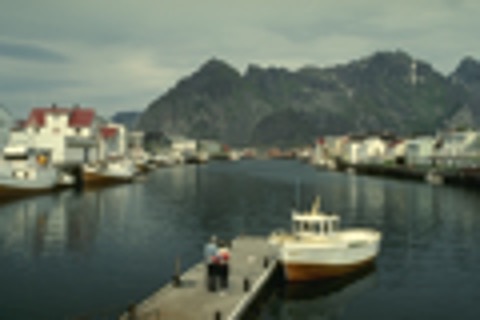}&
     	\includegraphics[width=1.5in]{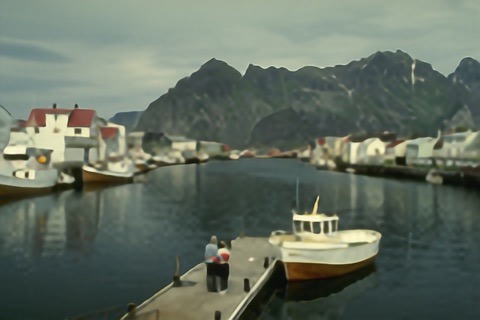} &
     	\includegraphics[width=1.5in]{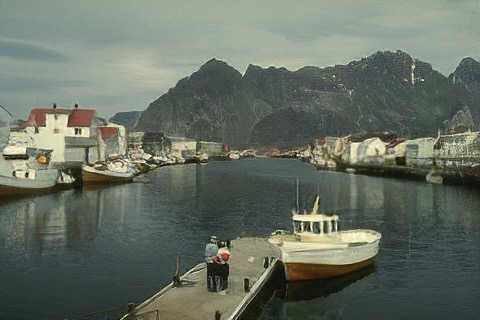} &
     	\includegraphics[width=1.5in]{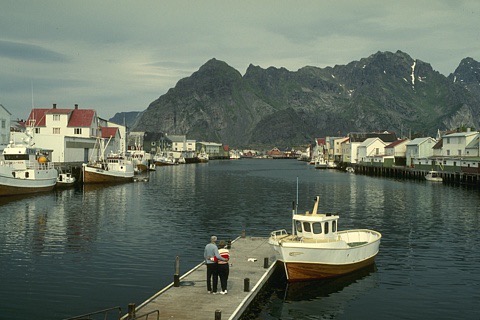} \\     	
     	\includegraphics[width=1.5in]{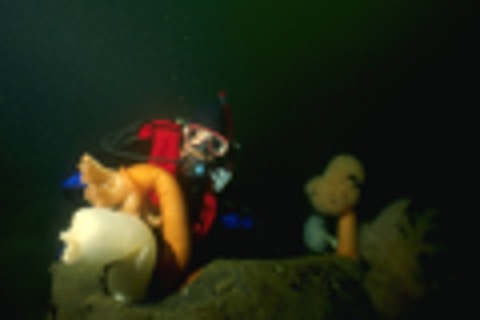}&
     	\includegraphics[width=1.5in]{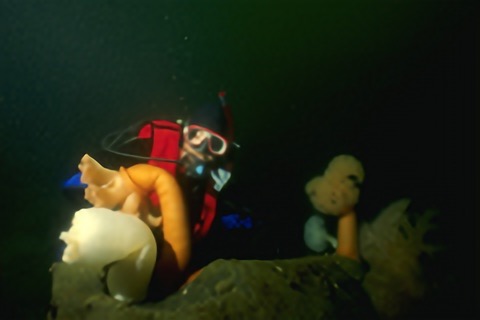} &
     	\includegraphics[width=1.5in]{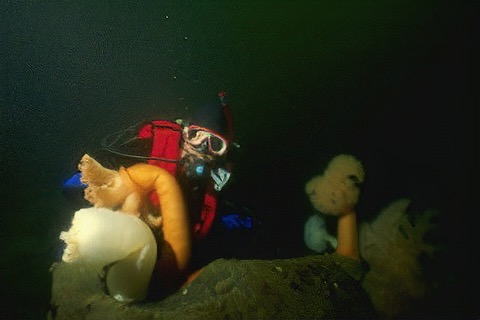} &
     	\includegraphics[width=1.5in]{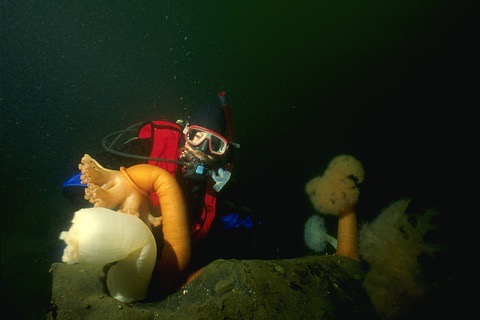} \\     	
     	\includegraphics[width=1.5in]{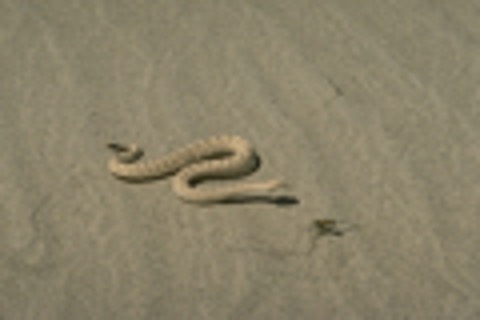}&
     	\includegraphics[width=1.5in]{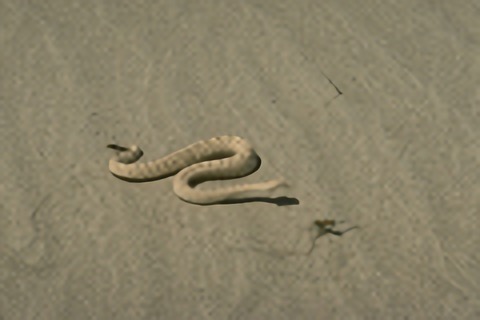} &
     	\includegraphics[width=1.5in]{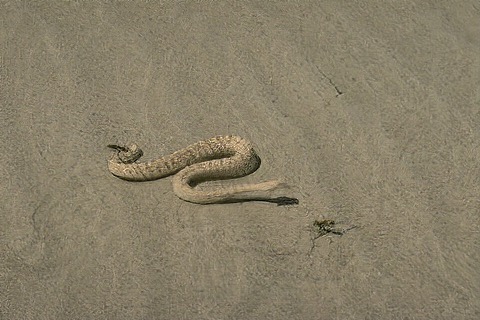} &
     	\includegraphics[width=1.5in]{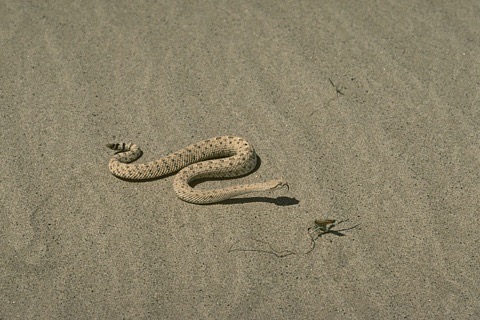} \\     	
     	\includegraphics[width=1.5in]{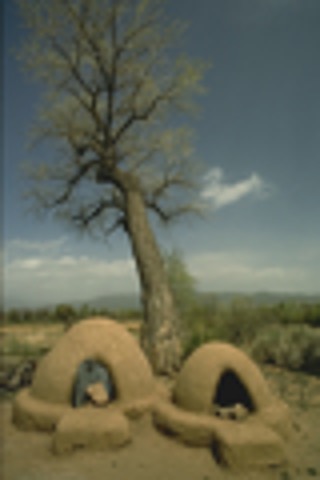}&
     	\includegraphics[width=1.5in]{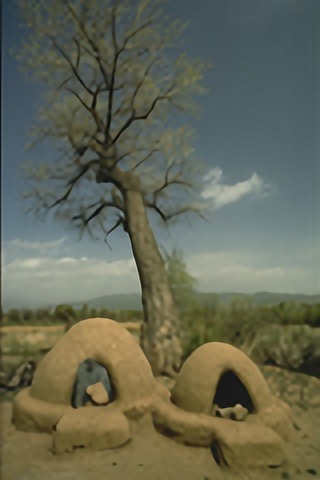} &
     	\includegraphics[width=1.5in]{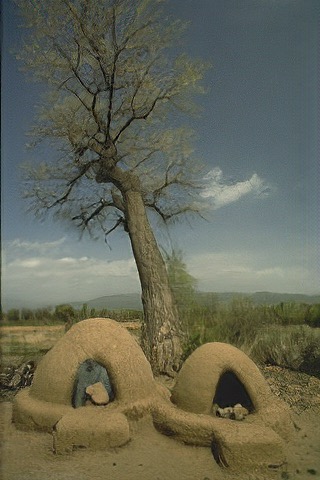} &
     	\includegraphics[width=1.5in]{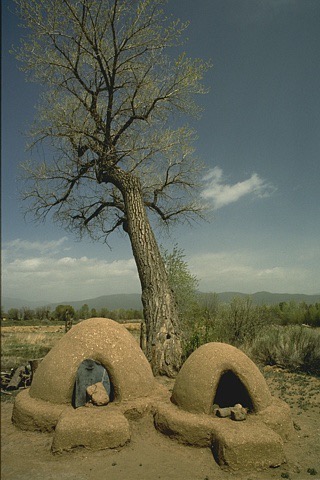} \\     	
     	\includegraphics[width=1.5in]{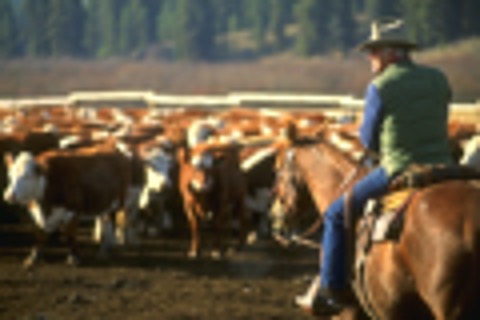}&
     	\includegraphics[width=1.5in]{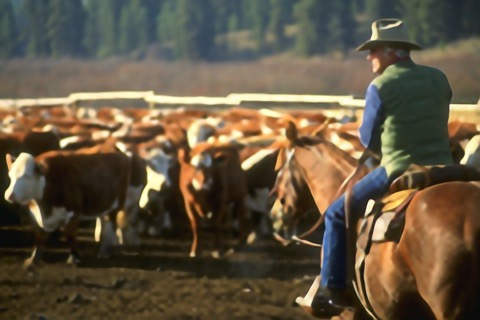} &
     	\includegraphics[width=1.5in]{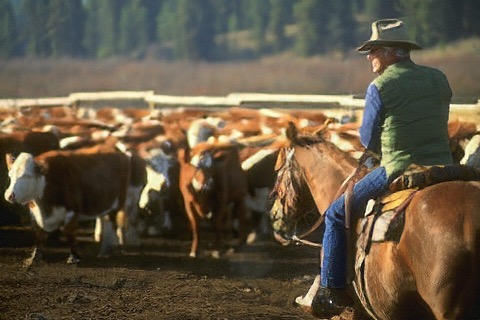} &
     	\includegraphics[width=1.5in]{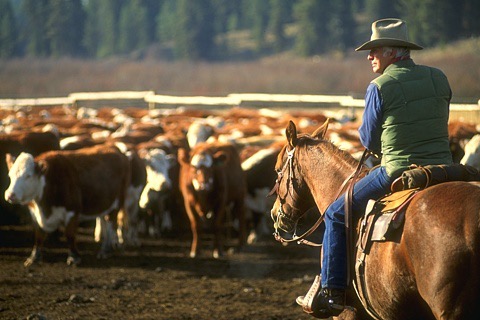} \\     	
  	\end{tabular}
  	\label{fig:app_BSD100}
  	\caption{Results for five random samples of \textbf{BSD100} using bicubic interpolation, SRResNet and SRGAN. [$4\times$ upscaling]} 
\end{figure*}

\end{document}